%% file: main.tex
\documentclass[runningheads]{llncs}
\usepackage{graphicx}
\usepackage{xcolor}
\usepackage{amsmath}
\usepackage{amsfonts}
\usepackage{amssymb}
\usepackage{mathabx}
\usepackage{xifthen}
\usepackage{subfig}
\usepackage[moderate]{savetrees}
\usepackage{times}
\DeclareMathOperator*{\argmax}{arg\,max}
\usepackage{pgf,tikz,pgfplots}
\usetikzlibrary{calc,patterns,decorations.pathmorphing,decorations.markings}
\usepackage{multirow}
\usepackage{adjustbox}
\begin{document}
\title{Causal Temporal Reasoning for\\Markov Decision Processes}
 \author{Milad Kazemi \and
 Nicola Paoletti}
 \authorrunning{M. Kazemi and N. Paoletti}

 \institute{Department of Informatics, King's College London}

\input{macro}
\addtolength{\parskip}{-0.64pt}

\maketitle              %
\input{abstract}
\input{intro}
\section{Background}\label{sec:background}
\input{scm}

\input{mdps}

\input{scm_mdp}
\input{causal_TL}
\input{procedures}

\input{experiments}

\input{related}
\input{conclusion}

\newpage
\bibliographystyle{splncs04}
\bibliography{refs}

\input{appendix}

\input{procedures_long}

\input{sanity_check}

\input{bisimilarity}

\end{document}

%% file: macro.tex
\newcommand{\np}[1]{{#1}}
\newcommand{\mk}[1]{{#1}}

\newcommand{\until}[1]{\mathcal{U}_{#1}}
\newcommand{\finally}[1]{\mathcal{F}_{#1}}
\newcommand{\globally}[1]{\mathcal{G}_{#1}}
\newcommand{\pinter}[3]{P_{#1}(#2\mid \textit{do}({#3}))}
\newcommand{\pcf}[4]{P_{#1}(#2_{#3},#4)}

\newcommand{\scm}{\mathcal{M}}
\newcommand{\ndovars}{\mathbf{V}}
\newcommand{\exovars}{\mathbf{U}}

\newcommand{\mdp}{\mathcal{P}}
\newcommand{\states}{\mathcal{S}}
\newcommand{\actions}{\mathcal{A}}

\newcommand{\probs}[1][]{%
\ifthenelse{\equal{#1}{}}{P_{\mathcal{P}}}{P_{\mathcal{P}(#1)}}%
}

\newcommand{\gumbelscm}[1]{\mathcal{M}^{#1}}

\definecolor{highlightcol}{HTML}{fcf9d1}
\definecolor{highlightcol1}{HTML}{d1c964}

\newcommand{\indep}{\rotatebox[origin=c]{90}{$\models$}}

\newcommand{\review}[1]{#1}

%% file: abstract.tex
\begin{abstract}
We introduce \textit{PCFTL (Probabilistic CounterFactual Temporal Logic)}, a new probabilistic temporal logic for the verification of Markov Decision Processes (MDP). PCFTL is the first to include operators for causal reasoning, allowing us to express interventional and counterfactual queries. Given a path formula $\phi$, an interventional property is concerned with the satisfaction probability of $\phi$ if we apply a particular change $I$ to the MDP (e.g., switching to a different policy); a counterfactual allows us to compute, given an observed MDP path $\tau$, what the outcome of $\phi$ would have been had we applied $I$ in the past. For its ability to reason about \textit{what-if} scenarios involving different configurations of the MDP, our approach represents a departure from existing probabilistic temporal logics that can only reason about a fixed system configuration. From a syntactic viewpoint, we introduce a generalized counterfactual operator that subsumes both interventional and counterfactual probabilities as well as the traditional probabilistic operator found in e.g., PCTL. From a semantics viewpoint, our logic is interpreted over a structural causal model  translation of the MDP, which gives us a representation amenable to counterfactual reasoning. We evaluate PCFTL in the context of safe reinforcement learning using a benchmark of grid-world models. 
\end{abstract}

%% file: intro.tex
\section{Introduction}

Temporal logic (TL) is arguably the primary language for the formal specification and reasoning about system correctness and safety. 
It has been successfully applied to the analysis of a wide range of systems, including programs~\cite{manna2012temporal}, cyber-physical systems~\cite{bartocci2018specification}, and stochastic models~\cite{kwiatkowska2007stochastic}. 
A limitation of existing TLs is that TL specifications are evaluated on a fixed configuration of the system, e.g., a fixed choice of control policy, communication protocol, or system dynamics. That is, they cannot express queries like \textit{``will the system remain safe if we switch to a different, high-performance controller?''}, or \textit{``what is the probability of failure if we had applied a different policy in the past?''} This kind of reasoning about different system conditions falls under the realm of causal inference~\cite{pearl_2009}, by which the first query is called an \textit{intervention} and the second a \textit{counterfactual}. 
Even though causal inference and TL-based verification are very well-established on their own, their combination hasn't been sufficiently explored in past literature (see Section~\ref{sec:related} for a more complete account of the related work). With this paper, we aim to bridge these two fields. 

We introduce \textit{PCFTL (Probabilistic CounterFactual Temporal Logic)}, 
the first probabilistic temporal logic that explicitly includes causal operators to express interventional properties (\textit{``what will happen if\ldots''}), counterfactual properties (\textit{``what would have happened if\ldots''}), and so-called \textit{causal effects}, defined as the difference of interventional or counterfactual probabilities between two different configurations. In particular, in this paper we focus on the analysis of \textit{Markov Decision Processes (MDPs)}, one of the main models in reinforcement learning, planning, and probabilistic verification. For MDPs, arguably the most relevant kind of causal reasoning concerns evaluating how a change in the MDP policy affects some outcome.  The outcome of interest for us is the satisfaction probability of a temporal-logic formula.

Interventions are ``forward-looking''~\cite{oberst2019counterfactual}, as they allow us to evaluate the probability of a TL property $\phi$ after applying a particular change $X\gets X'$ to the system. Counterfactuals are instead ``retrospective''~\cite{oberst2019counterfactual}, telling us what might have happened under a different condition: having observed an MDP path $\tau$, they allow us to evaluate $\phi$ on the \textit{what-if} version of $\tau$, i.e., the path that we would have observed if we had applied $X\gets X'$ at some point in the past. 
Causal effects~\cite{guo2020survey} allow us to establish the impact of a given change at the level of the individual path or overall, and they quantify the increase in the probability of $\phi$ induced by a manipulation $X\gets X'$. 

Causal and counterfactual reasoning has gained a lot of attention in recent years due to its power in observational data studies: with counterfactuals, one can answer \emph{what-if} questions relative to an observed path, i.e., without having to intervene on the real system (which might jeopardize safety) but using observational data only. Our PCFTL logic enables this kind of reasoning in the context of formal verification. 

Our approach to incorporating causal inference in temporal logic involves only a minimal extension of traditional probabilistic logics: PCFTL is a bounded-horizon variant of $\text{PCTL}^{\star}$~\cite{hansson1994logic,baier1997symbolic,bertrand2012bounded} where the probabilistic operator $P_{\bowtie p}(\phi)$, which checks whether the probability of $\phi$ satisfies threshold $\bowtie p$, is replaced with a generalized counterfactual operator $I_{@ t}.P_{\bowtie p}(\phi)$, which concerns the probability of $\phi$ if we had applied intervention $I$ at $t$ time steps in the past. Albeit minimal, such an extension provides great expressive power: if $t>0$, then the operator corresponds to a counterfactual query; if $t=0$, it derives instead an interventional probability; if both $t=0$ and $I$ is empty, then it corresponds to the traditional $P_{\bowtie p}(\phi)$ operator. 
Importantly, our operator generalizes the usual notion of counterfactuals as it can also reason about the counterfactual future beyond the observed path $\tau$ (when $t>0$ and the time bounds in $\phi$ extend beyond the length of $\tau$), while the usual notion focuses only on the observed past.

\review{Consider an example of robot navigation, where $\phi$ is a bounded reach-avoid specification (the robot must reach a goal by some deadline while avoiding obstacles), and the intervention $I = (\pi \gets \pi')$ is one where the nominal robot policy $\pi$ is changed into a different navigation policy $\pi'$.  Suppose we observe a path $\tau$ of the robot (under $\pi$). Then, the interventional query $I_{@ 0}.P_{\bowtie p}(\phi)$ allows us to evaluate the reach-avoid probability after we change the robot's policy (by applying $I$) at the current robot's state (i.e., at the last state of $\tau$). On the other hand, the counterfactual query $I_{@ 5}.P_{\bowtie p}(\phi)$ allows us to evaluate the reach-avoid probability in a what-if version of $\tau$ where the policy was changed at $5$ time steps in the past (i.e., at the sixth-last state of $\tau$). 
 This counterfactual analysis is useful when, for instance, the observed path $\tau$ is rare and uncommon, and we want to compare the outcomes of $\pi'$ and  $\pi$ under the same circumstances that generated $\tau$. Due to the rarity of the observed path, we cannot reliably retrieve such a hypothetical $\tau$ by performing Monte-Carlo simulations of the robot under $\pi'$; we need counterfactuals. Or, if $\tau$ violates $\phi$, we might want to verify that such a violation wouldn't have happened under the same circumstances if $\pi'$ (rather than $\pi$) was in place.}

Unlike existing logics, PCFTL formulas are interpreted with respect to an observed MDP path $\tau$ (rather than a single MDP state) and a so-called \textit{structural causal model (SCM)} translation of the MDP~\cite{pearl_2009,glymour2016causal,oberst2019counterfactual} (rather than the MDP itself). This translation is necessary because SCMs have a particular form that facilitates the computation of interventional and counterfactual distributions. 
By relying on efficient statistical model checking procedures, we evaluate PCFTL on a reinforcement learning benchmark~\cite{minigrid} involving multiple 2D grid-world environments, goal-oriented tasks, and interventional and counterfactual properties under different neural-network policies. These results demonstrate the usefulness of PCFTL in AI safety, but our approach could enhance the verification of probabilistic models in a variety of domains, from distributed systems to security and biology. 

%% file: scm.tex
\subsection{Causal Reasoning with Structural Causal Models}\label{sec:scms}
Structural Causal Models (SCMs)~\cite{pearl_2009,glymour2016causal,guo2020survey} are equation-based models to specify and reason about causal relationships involving some variables of interest. 
\begin{definition}[Structural Causal Model (SCM)]\label{def:scm}
An SCM is a tuple $\scm=(\exovars,\ndovars,\mathcal{F},P(\exovars))$ where
$\exovars$ is a set of (mutually independent) \emph{exogenous variables}, with $P(\exovars)=\bigtimes_{U \in \exovars} P(U)$ being the distribution of $\exovars$; $\ndovars$ is a set of \emph{endogenous variables}; the value of each $V \in \ndovars$ is determined by a function $V = f_V(\mathbf{PA}_V, U_V)$ where $\mathbf{PA}_V \subseteq \ndovars$ is the set of direct causes of $V$ and $U_V\in \exovars$; and $\mathcal{F}$ is the set of functions $\{f_V\}_{V \in \ndovars}$. Assignments in $\mathcal{F}$ are such that the corresponding \emph{causal graph}  $(\ndovars \cup \exovars, \{ (U_V,V) \mid V \in \ndovars \}\cup \{(V',V) \mid V \in \ndovars,  V' \in \mathbf{PA}_V \})$ is acyclic. 
\end{definition}

With acyclic assignments, we ensure that each variable cannot be a direct or indirect cause of itself. Also, we require that each exo-variable cannot affect more than one endo-variable, i.e., for every $V_1,V_2 \in \ndovars$, $V_1\neq V_2 \iff U_{V_1} \neq U_{V_2}$\footnote{This (common) assumption rules out any source of unobserved confounding.}.  
In an SCM, the values of the exo-variables $\exovars$ are determined by factors outside the model, which is modelled by some distribution $P(\exovars)$. 
Exo-variables are \textit{unobserved} variables which act as the source of randomness in the system.  Indeed, for a fixed realization $\mathbf{u}$ of $\exovars$, i.e., a concrete unfolding of the system's randomness, the values of $\ndovars$ become deterministic, as they are uniquely determined by $\mathbf{u}$ and the causal processes $\mathcal{F}$. A concrete value $\mathbf{u}$ of $\exovars$ is also called \textit{context} (or unit). We denote by $P_{\scm}(\ndovars)$ the so-called \textit{observational distribution} of $\ndovars$, that is, the data-generating distribution entailed by the SCM $\mathcal{F}$ and $P(\exovars)$. Also, we denote with $V(\mathbf{u})$ the value of $V$ uniquely determined by context $\mathbf{u}$.

\paragraph{Interventions.} With SCMs, one can establish the causal effect of some input variable $X$ on some output variable $Y$ by evaluating $Y$ after ``forcing'' some specific values $x$ on $X$, an operation called \textit{intervention}. 
Applying $X\gets x$ means replacing the RHS of $X = f_X(\mathbf{PA}_X, U_X)$ with $x$. 
Interventions allow to establish the true causal effect of $X$ on $Y$ by comparing the so-called \textit{post-interventional distribution} $P_{\scm[X\gets x]}(Y)$ at different values $x$,  where $\scm[X\gets x]$ is the SCM obtained from $\scm$ by applying $X\gets x$. By ``disconnecting'' $X$ from any of its possible causes, interventions prevent any source of spurious association between $X$ and $Y$~\cite{glymour2016causal} (i.e., caused by variables other than $X$ and that are not descendants of $X$)\footnote{Note that $P_{\scm}(Y \mid X=x)$ is, in general, different from the desired $P_{\scm[X\gets x]}(Y)$ because conditioning on $X=x$ alone doesn't prevent unwanted spurious associations.}. 
In the following we will use the notation $I$ (and $\scm[I]$) to denote a set of interventions $I=\{V_i \gets v_i\}_i$.

\paragraph{Counterfactuals.} 
On observing a particular realization $\mathbf{v}$ of the SCM variables $\ndovars$, counterfactuals answer the following question: \textit{what would have been the value of some variable $Y$ for observation $\mathbf{v}$ if we had applied intervention $I$ on our model $\scm$?} This corresponds to evaluating $\ndovars$ in a hypothetical world characterized by the same context (i.e., same realization of random factors) that generated the observation $\mathbf{v}$ but under a different causal process. 

There are three steps to compute counterfactuals~\cite{glymour2016causal}: 
1)~\textit{abduction}, where we estimate the context given the observation, i.e., derive $P(\exovars \mid \ndovars=\mathbf{v})$;
2)~\textit{action}, where we modify the SCM by applying some intervention $I$; and 
3)~\textit{prediction}, where we evaluate $\ndovars$ under the manipulated model $\scm[I]$ and the inferred context. 

We denote by $\scm(\mathbf{v})[I]$
the \textit{counterfactual model} obtained by replacing $P(\exovars)$ with $P(\exovars \mid \ndovars=\mathbf{v})$ in the SCM $\scm$ and then applying intervention $I$. Note that here $\mathbf{v}$ is a realization of $\ndovars$ under $\scm$ and not under $\scm[I]$.

As explained above, each observation $\ndovars=\mathbf{v}$ can be seen as a deterministic function of a particular value $\mathbf{u}$ of $\exovars$. Therefore, the counterfactual model is deterministic too, assuming that such $\mathbf{u}$ can be identified from $\ndovars=\mathbf{v}$. 
However, inferring $\mathbf{u}$ precisely is often not possible (as discussed later), resulting in a (non-Dirac) posterior distribution of contexts $P(\exovars \mid \ndovars=\mathbf{v})$ and thus, a stochastic counterfactual value.

\begin{remark}\label{rem:abduction}
To compute the counterfactual distribution $P_{\scm(\mathbf{v})[I]}(Y)$ of some $Y \in \ndovars$, the  abduction step is not necessary for all the exo-variables, but only for those $U_V$ where $V$ appears in a causal path between the variables manipulated in $I$ and $Y$. 
In particular, $Y$ remains equal to its value in $\mathbf{v}$, denoted $Y=\mathbf{v}_Y$, 
for all $Y$ that are not causal descendants of the variables manipulated by $I$. Thus, if $I=\emptyset$, then $Y=\mathbf{v}_Y$ for all $Y$. Also, to infer $P(U_{V} \mid \ndovars=\mathbf{v})$ we only require  $\mathbf{v}_{V}$ and $\mathbf{v}_{V'}$ for all $V'\in \mathbf{PA}_V$.
\end{remark}

\subsubsection{Causal Effects.} Estimating a causal effect amounts to comparing some variable $Y$ (outcome, output) under different values of some other variable $X$ (treatment, input). 
Interventions and counterfactuals enable this task by ruling out spurious association between $X$ and $Y$, as discussed above. There are three main estimators of causal effects:

\paragraph{Individual Treatment Effect (ITE).} For a context $\mathbf{u}$, the ITE of $Y\in \ndovars$ between interventions $I_1$ and $I_0$ is defined as $Y_{I_1}(\mathbf{u})-Y_{I_0}(\mathbf{u})$, 
where $Y_{I_i}(\mathbf{u})$ is the counterfactual value of $Y$ induced by $\mathbf{u}$ under the post-intervention model $\scm[I_i]$. As explained above, we don't have direct access to the exogenous values $\mathbf{u}$ but only to realizations $\mathbf{v} \sim P_{\scm}(\ndovars)$. Thus, below we define the ITE as a function of $\mathbf{v}$ (rather than $\mathbf{u}$) by plugging in the average counterfactual value of $Y$ w.r.t.\ the posterior $P(\exovars \mid \ndovars=\mathbf{v})$: 
    \begin{equation}\label{eq:ite2}
    \textit{ITE}(Y,I_1,I_0,\mathbf{v}) = \mathbb{E}_{\scm(\mathbf{v})[I_1]}[Y] - \mathbb{E}_{\scm(\mathbf{v})[I_0]}[Y].
    \end{equation}
\paragraph{Average Treatment Effect (ATE).} ATE is used to estimate causal effects at the population level and is defined as the expected value (w.r.t.\ $P(\exovars)$) of the individual treatment effect, or equivalently, as the difference of post-interventional expectations: 
    \begin{equation}
    \textit{ATE}(Y,I_1,I_0)=\mathbb{E}_{\scm[I_1]}[Y] - \mathbb{E}_{\scm[I_0]}[Y].
    \end{equation}
\paragraph{Conditional Average Treatment Effect (CATE).} The CATE is the conditional version of ATE. This estimator is useful when the treatment effect may vary across the population depending on the value of some variables $V$:
    \begin{equation}
    \textit{CATE}(Y,I_1,I_0,v)=\mathbb{E}_{\scm[I_1]}[Y\mid V=v] - \mathbb{E}_{\scm[I_0]}[Y\mid V=v].
    \end{equation}

%% file: mdps.tex
\subsection{Markov Decision Processes (MDPs)}
MDPs are a class of stochastic models to describe sequential decision making processes, where at each step $t$, an agent in state $s_i$ performs some action $a_i$ determined by some policy $\pi$ ending up in state $s_{i+1}\sim P(s\mid s_i, a_i)$. The agent receives some reward $R(s_i,a_i)$ for performing $a_i$ at $s_i$. Here we  focus on MDPs with finite state and action spaces and deterministic policies. Moreover, each MDP state satisfies a (possibly empty) set of atomic propositions, with $\mathit{AP}$ being the set of atomic propositions.

\begin{definition}[Markov Decision Process (MDP)]
An MDP is a tuple $\mdp=(\states,\actions,\probs,P_I,R,L)$ where $\states$ is the \emph{state space}, $\actions$ is the set of \emph{actions}, $\probs: (\states \times \actions \times \states) \rightarrow [0,1]$ is the \emph{transition probability} function, $P_I:\states \rightarrow [0,1]$ is the \textit{initial state distribution}, $R: (\states \times \actions) \rightarrow \mathbb{R}$ is the \emph{reward function}, and $L:\states\rightarrow 2^{\mathit{AP}}$ is a \emph{labelling function}, which assigns to each state $s\in \states$ the set of atomic propositions that are valid in $s$. A (deterministic) \emph{policy} $\pi$ for $\mdp$ is a function $\pi:\states\rightarrow\actions$.
\end{definition}
For a state $s\in \states$, we denote with $\mdp(s)$ the equivalent of $\mdp$ where $P_I(s)=1$, i.e., where $s$ is the only possible initial state.

\begin{definition}[MDP path]\label{def:mdp_path}
A path $\tau=(s_1,a_1),(s_2,a_2),\ldots$ of an MDP
$\mdp=(\states,\actions,\probs,P_I,R,L)$
induced by a policy $\pi$ is a sequence of state-action pairs where $s_i\in \states$ and $a_i=\pi(s_i)$ for all $i\geq 1$. 
The probability of $\tau$ is given by $\probs(\tau)=P_I(s_1)\cdot \prod_{i\geq 1} \probs(s_{i+1}\mid s_i, a_i)$. For a finite path $\tau=(s_1,a_1),\ldots,(s_k,a_k)$, 
we denote by $\mathit{Paths}_{\pi}(\tau)$
the set of all (infinite) paths with prefix $\tau$ induced by policy $\pi$, which has probability $\probs(\mathit{Paths}_{\pi}(\tau))=\probs(\tau)$. 
\end{definition}

We denote by $|\tau|$ the length of the path, by $\tau[i]$ the $i$-th element of $\tau$ (for $0<i\leq |\tau|$), by $\tau[i:]$ the suffix of $\tau$ starting at position $i$ (inclusive), and by $\tau[i:i+j]$ the subsequence spanning positions $i$ to ${i+j}$ (inclusive). When $\tau$ is finite, we also use negative indices $\tau[-i]$ to denote $\tau[|\tau|-i]$. Hence, $\tau[0]=\tau[|\tau|]$ corresponds to the last element of $\tau$. Even though $\tau[i]$ denotes the pair $(s_i,a_i)$ of the path, we will often use it, when the context is clear, to denote only the state $s_i$. 
We slightly abuse notation and write $\mathit{Paths}_{\pi}(s)$ to denote the set of paths induced by $\pi$ and starting with $s$.

%% file: scm_mdp.tex
\subsection{SCM-based encoding of MDPs}\label{sec:scm_mdp_encoding}
We now present the SCM-based encoding of MDPs introduced in~\cite{oberst2019counterfactual}. For a given path length $T$, the SCM $\gumbelscm{\mdp,\pi,T}$ induced by an MDP $\mdp$ and a policy $\pi$ characterizes the unrolling of paths of $\mdp$ of length $T$, i.e., it has endo-variables $S_t$ and $A_t$ describing the MDP's state and action at each $t=1,\ldots,T$ which are defined by:
\begin{equation}\label{eq:mdp_scm}
    S_{t+1} = f(S_{t},A_{t}, U_{t}); \quad A_t = \pi(S_t); 
    \quad S_1 = f_0(U_0),
\end{equation}
where the probabilistic state transition at $t$, $\probs(S_{t+1} \mid S_t, A_t)$, is encoded as a deterministic function $f$ of $S_{t}$, $A_{t}$, and the (random) exo-variable $U_t$, while the random choice of the initial state, $P_I(S_1)$, as a deterministic function $f_0$ of the exo-variable $U_0$. 

\begin{figure}
  \begin{minipage}[b]{0.5\textwidth}
    \includegraphics[width=\linewidth]{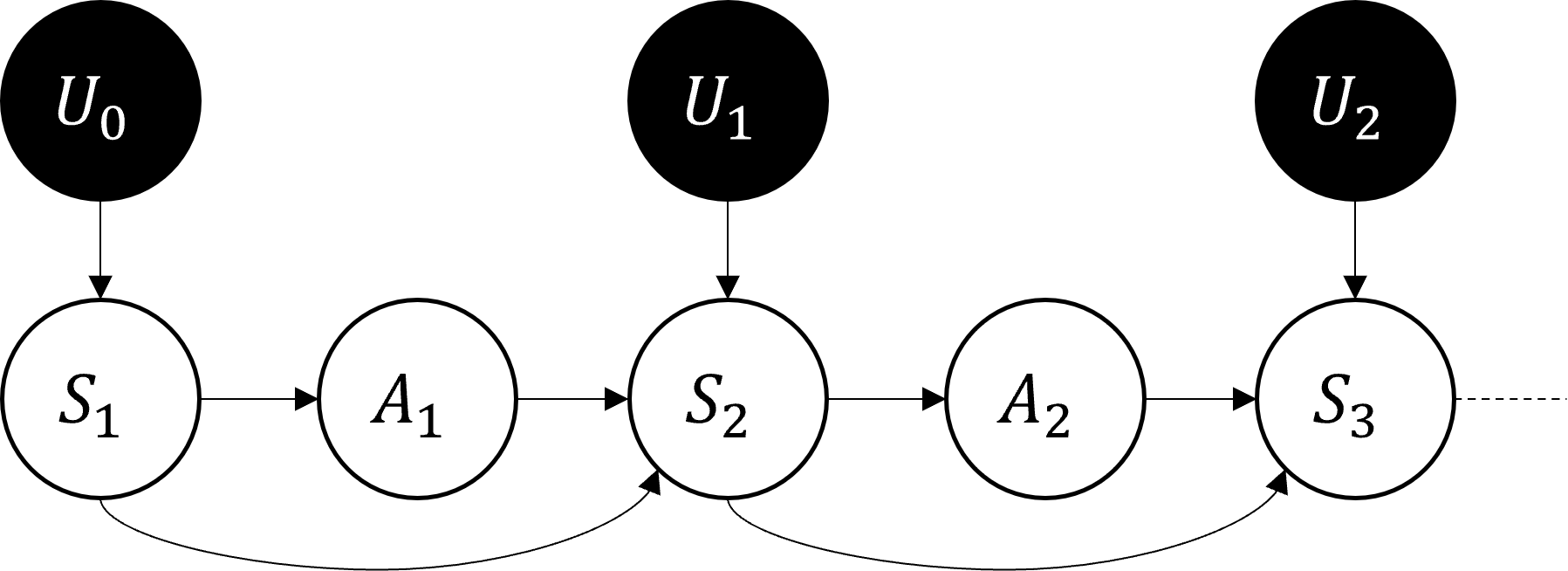}
  \end{minipage}\hfill
  \begin{minipage}[b]{0.45\textwidth}
    \caption{
       Causal diagram for the SCM encoding of an MDP. Black circles represents exogenous variables, white circles endogenous ones.
    }     \label{fig:mdp_scm}
  \end{minipage}
\end{figure}

\mk{We stress that the SCM encoding does not require any assumptions about the structure of the MDP: such encoding results in an acyclic graph, while the original MDP need not be.}
Figure~\ref{fig:mdp_scm} shows the causal diagram resulting from this SCM encoding.

Note that both $\probs(S_{t+1} \mid S_t, A_t)$ and $P_I(S_1)$ are categorical distributions and encoding them in the above SCM form (i.e., as functions of a random variable) is not obvious. Oberst and Sontag~\cite{oberst2019counterfactual} have recently proposed 
a solution called 
\textit{Gumbel-Max SCM}, given by 
\begin{equation}\label{eq:gumbel-max-scm}
    S_{t+1} = f(S_{t},A_{t}, U_{t}=(G_{s,t})_{s\in \states}) = \argmax_{s\in \states}\left\{\log\left(\probs(S_{t+1}= s \mid S_t, A_t)+ G_{s,t}\right)\right\}, 
\end{equation}
where, for $s\in \states$ and $t\in 1=\ldots,T$, $G_{s,t}\sim \mathrm{Gumbel}$. This is based on the Gumbel-Max trick, by which one can sample from a categorical distribution with $k$ categories (corresponding to the $|S|$ MDP states in our case) by first drawing realizations $g_1,\ldots,g_k$ of a standard Gumbel distribution and then by setting the outcome to $\argmax_{j}\left\{\log\left(P(Y=j)+ g_j\right)\right\}$. The Gumbel sample $g_1,\ldots,g_k$ is obtained by drawing $u_1,\ldots,u_k \sim_{iid} \mathrm{Unif}(0,1)$ and setting $g_i=-\log(-\log (u_i))$. Importantly, the \textit{Gumbel-Max SCM} encoding enjoys a desirable property called \textit{counterfactual stability} (see  Appendix~\ref{app:cf_stab} for more details)\footnote{In contrast, the approach based on the inverse CDF trick, where $f(S_{t},A_{t}, U_{t})$ is the $U_t$-quantile of $\probs(S_{t+1} \mid S_t, A_t)$ and $U_{t}\sim \mathrm{Unif}(0,1)$, does not enjoy counterfactual stability and is highly sensitive to permutations of the state ordering (assuming some ordering is required by the quantile function).}.

\begin{definition}[Counterfactual stability~\cite{oberst2019counterfactual}]
An SCM $\scm$ satisfies \emph{counterfactual stability} relative to a categorical variable $Y$ of $\scm$ if whenever we observe $Y=i$ under some intervention $I$, then the counterfactual value of $Y$ under $I'\neq I$ remains $Y=i$ unless $I'$ increases the relative likelihood of an alternative outcome $j\neq i$, i.e., unless  ${P_{\scm[I']}(Y=j)}/{P_{\scm[I]}(Y=j)}>{P_{\scm[I']}(Y=i)}/{P_{\scm[I]}(Y=i)}$. 
\end{definition}

\begin{proposition}[Correctness of the Gumbel-Max encoding]\label{prop:encoding}
Given an MDP $\mdp$, policy $\pi$, time bound $T$, then for any path $\tau$ of $\mdp$ induced by $\pi$ of length $T$, we have  $P_{\gumbelscm{\mdp,\pi,T}}(\tau)=\probs(\tau)$, where $\gumbelscm{\mdp,\pi,T}$ is the Gumbel-Max SCM for $\mdp$, $\pi$, and $T$.
\end{proposition}
\begin{proof}
This readily follows from the fact that, by the Gumbel-Max trick, the assignment $S_{t+1} = f(S_{t},A_{t}, (G_{s,t})_{s\in \states})$ in~\eqref{eq:gumbel-max-scm} is equivalent to sampling $S_{t+1} \sim \probs(S \mid S_t, A_t)$.
\end{proof}
We will use $\mathit{Paths}_{\gumbelscm{\mdp,\pi,T}}(\tau)$ to denote the set of paths of length $T$ with prefix $\tau$ admitted by the Gumbel-Max SCM $\gumbelscm{\mdp,\pi,T}$, i.e., such that $P_{\gumbelscm{\mdp,\pi,T}}(\tau')>0$ for all $\tau' \in \mathit{Paths}_{\gumbelscm{\mdp,\pi,T}}(\tau)$. 

\paragraph{Counterfactual inference.} 

Given we observed an MDP path $\tau=(s_1,a_1),\ldots,(s_T,a_T)$, counterfactual inference in this setting entails deriving $P((G_{s,t})_{s\in \states}^{t=0,\ldots,T-1}\mid \tau)$, that is, finding values for the Gumbel exo-variables compatible with  $\tau$. By the Markov property, the above can be factorized as follows:
$$P((G_{s,t})_{s\in \states}^{t=0,\ldots,T-1}\mid \tau)=P((G_{s,0})_{s\in \states} \mid s_1)\cdot\prod_{t=1}^{T-1} P((G_{s,t})_{s\in \states}\mid s_t,a_t,s_{t+1}).$$ 
However, the mechanism of~\eqref{eq:gumbel-max-scm} is non-invertible, i.e., given $s_t$ and $a_t$, there might be multiple values of $(G_{s,t})_{s\in \states}$ leading to the same $s_{t+1}$. This implies that \textit{MDP counterfactuals can't be uniquely identified}, a problem that affects categorical counterfactuals in general and not just Gumbel-Max SCMs~\cite{oberst2019counterfactual}. 

As pointed out in~\cite{oberst2019counterfactual}, we can still perform (approximate) posterior inference of $P((G_{s,t})_{s\in \states} \mid s_t,a_t,s_{t+1})$ by \textit{rejection sampling}: we first sample from the prior $P((G_{s,t})_{s\in \states})$ and then reject all the Gumbel realizations $(g_{s,t})_{s\in \states}$ for which $f(s_t,a_t,(g_{s,t})_{s\in \states})\neq s_{t+1}$, i.e.,  $(g_{s,t})_{s\in \states}$ is incompatible with the observed MDP step. 

%% file: causal_TL.tex
\section{PCFTL: a Probabilistic Temporal Logic with Interventions, Counterfactuals and Causal Effects}\label{sec:ptl_cf}

We introduce a probabilistic temporal logic for MDPs called \textit{PCFTL (Probabilistic CounterFactual Temporal Logic)}, the first logic of its kind to enable reasoning about interventions, counterfactuals and causal effects. A formula of this logic is interpreted over a structural causal model, and in particular, a Gumbel-Max SCM $\gumbelscm{\mdp,\pi,T}$ encoding paths of length $T$ of an MDP $\mdp$ under policy $\pi$. 
\review{
PCFTL is an extension of bounded $\text{PCTL}^{\star}$ ($\text{BPCTL}^{\star}$), and so we now introduce an SCM-based semantics for $\text{BPCTL}^{\star}$ as a preliminary step.
}

\paragraph{\mk{$\text{BPCTL}^{\star}$} and SCM-based semantics.} The syntax of \mk{$\text{BPCTL}^{\star}$} is as follows\mk{~\cite{hansson1994logic,baier1997symbolic,bertrand2012bounded}}:
\begin{align*}
    \Phi ::=  \top \mid \mk{\rho} \mid \neg \Phi \mid \Phi \wedge \Phi 
    \mid  P_{\bowtie p}(\phi) \mid R^{[a,b]}_{\bowtie r}; \qquad
    \phi ::=  \Phi \mid \neg \phi \mid \phi \wedge \phi \mid \phi \until{[a,b]} \phi    
\end{align*}
where $\mk{\rho} \in \mathit{AP}$, $\bowtie \in \{<,\leq,\geq,>\}$, $p\in[0,1]$, $r \in \mathbb{R}$, and $[a,b]$ is an interval in $\mathbb{Z}^{\geq 0}$. 

In our SCM-based semantics, a formula $\Phi$ is interpreted over a pair $(\gumbelscm{\mdp,\pi,T},\tau)$, where $\gumbelscm{\mdp,\pi,T}$ is the SCM encoding of MDP $\mdp$ under policy $\pi$ and time bound $T$, and $\tau$ is the path observed up to the current time. Clearly, by the Markov property, to evaluate $\Phi$ it is sufficient to know only $\tau[0]$, i.e., the last state of $\tau$. However, keeping track of the history of states in $\tau$ will become useful later in our definition of counterfactual operators. 

The probabilistic formula $P_{\bowtie p}(\phi)$ is true if the set of paths of $\gumbelscm{\mdp,\pi,T}$ starting from the last state of $\tau$ have probability consistent with the bound $\bowtie p$. The argument $\phi$ of $P_{\bowtie p}$ is a bounded temporal logic formula, aka \textit{path formula}. Formula $\phi_1 \until{[a,b]} \phi_2$ is satisfied by paths where $\phi_2$ holds at some time point within interval $[a,b]$ and $\phi_1$ holds always before that point. Other standard bounded temporal operators are derived as: 
$\finally{[a,b]} \phi \equiv \top \until{[a,b]} \phi$ (\textit{eventually}),
$\globally{[a,b]} \phi \equiv \neg \finally{[a,b]} \neg \phi$ (\textit{always}), and
$\mathcal{X} \phi \equiv \finally{[1,1]} \phi$ (\textit{next}).
Formula $R^{[a,b]}_{\bowtie r}$ is true if the expected reward accumulated in interval $[a,b]$ satisfies $\bowtie r$. 

In the remainder of the paper, we assume that $T$ is selected to be long enough such that we can determine the satisfiability of any  $\phi$ expressions. Since we restrict to bounded temporal operators, we can always find such a $T<\infty$. Also, we will use the more compact notation $\gumbelscm{\mdp}$, keeping $\pi$ and $T$ implicit. 
We write $(\gumbelscm{\mdp},\tau)\models \Phi$ to say that $\Phi$ is satisfied by $\gumbelscm{\mdp}$ and $\tau$.  

\begin{definition}[SCM semantics of \mk{$\text{BPCTL}^{\star}$}]\label{def:ptl_sec}
Given a \mk{$\text{BPCTL}^{\star}$} formula $\Phi$, an SCM $\gumbelscm{\mdp}$ for an MDP $\mdp$, a path $\tau$ of $\gumbelscm{\mdp}$, with $s=\tau[0]$ being the last state of $\tau$, then the satisfaction relation $\models$ is defined below:
\[
\begin{array}{rlll}
    (\gumbelscm{\mdp},\tau) \models & \mk{\rho} & \text{ if }& \mk{\rho}\in L(s)\\
    (\gumbelscm{\mdp},\tau) \models & \neg \Phi & \text{ if }& (\gumbelscm{\mdp},\tau) \nvDash \Phi\\
    (\gumbelscm{\mdp},\tau) \models & \Phi_1 \wedge \Phi_2 & \text{ if }& ( (\gumbelscm{\mdp},\tau) \models \Phi_1) \wedge ( (\gumbelscm{\mdp},\tau) \models \Phi_2)\\
    
    (\gumbelscm{\mdp},\tau) \models & P_{\bowtie p}(\phi) & \text{ if } &
    P_{\gumbelscm{\mdp(s)}}(\{\tau' \in \textit{Paths}_{\gumbelscm{\mdp(s)}}(s) \mid (\gumbelscm{\mdp(s)}, \tau', 1) \models \phi \}) \bowtie p\\
    
    (\gumbelscm{\mdp},\tau)  \models & R^{[a,b]}_{\bowtie r} & \text{ if } & 
    \sum_{\tau' \in \textit{Paths}_{\gumbelscm{\mdp(s)}}(s)} \left(P_{\gumbelscm{\mdp(s)}}(\tau')\cdot \sum_{i=a}^b R(\tau'[i])\right) \bowtie r\\[10pt]
    
    (\gumbelscm{\mdp},\tau, t) \models & \Phi & \text{ if }& (\gumbelscm{\mdp}, \tau[1:t]) \models \Phi\\

    (\gumbelscm{\mdp},\tau, t) \models & \neg \phi & \text{ if }& (\gumbelscm{\mdp},\tau, t) \nvDash \phi\\
    (\gumbelscm{\mdp},\tau, t) \models & \phi_1 \wedge \phi_2 & \text{ if }& ( (\gumbelscm{\mdp},\tau, t) \models \phi_1) \wedge ( (\gumbelscm{\mdp},\tau, t) \models \phi_2)\\
    
    (\gumbelscm{\mdp}, \tau, t) \models & \phi_1 \until{[a,b]} \phi_2 & \text{ if }& \exists t_1 \in [a,b]. ( (\gumbelscm{\mdp}, \tau, t+t_1) \models \phi_2 \wedge \\ 
    & & & \hfill \forall t_2 \in [0,t_1) . ( (\gumbelscm{\mdp}, \tau, t+t_2) \models \phi_1 )).
\end{array}
\]
\end{definition}

The main difference between the above semantics and the usual (B)PCTL$^{\star}$ semantics is that in the latter, formulas are interpreted over a model-state pair $(\mdp,s)$. We instead keep a path $\tau$ of past observed states to enable counterfactual reasoning, as we will see next\footnote{For this reason, in determining the satisfaction of path formula $(\gumbelscm{\mdp},\tau, t) \models \Phi$, we carry over the path prefix $\tau[1:t]$, while in (B)PCTL$^{\star}$, $\Phi$ is evaluated over state $\tau[t]$.}. 

\review{
Nevertheless, the above SCM semantics remains equivalent to the usual one because: 1) the satisfaction of a (B)PCTL$^{\star}$ state formula depends only on the last state of the path, and 2) Proposition~\ref{prop:encoding} ensures that the probability measure over paths induced by the Gumbel-Max SCM (and used in our semantics) is equivalent to the probability measure over paths induced by the original Markov chain (used in the usual (B)PCTL$^{\star}$ semantics). 
}

\subsection{Syntax and Semantics of PCFTL}\label{sec:pcftl}
PCFTL extends \mk{$\text{BPCTL}^{\star}$} by introducing a \textit{counterfactual operator} $I_{@ t}.P_{\bowtie p}(\phi)$. This formula is true w.r.t.\ path $\tau$ if $P_{\bowtie p}(\phi)$ is true in the counterfactual model induced by observing $\tau$ and applying intervention $I$. In particular, we apply the intervention $I$ at $t$ steps back in the past, i.e., at state $s=\tau[-t]$, and continue applying $I$ thereafter. Then, the probability of $\phi$ is evaluated starting from state $s=\tau[-t]$, i.e., at the time of intervention, not from the last (counterfactual) state of the path (to do so, one can simply replace $\phi$ with $\finally{[t,t]}\phi$). PCFTL includes a \textit{counterfactual reward operator} $I_{@ t}.R^{[a,b]}_{\bowtie r}$ defined in a similar fashion, and two \textit{causal effect operators} as well, $\Delta_{@t}^{I_1,I_0}.P_{\bowtie p}(\phi)$ and $ \Delta_{@t}^{I_1,I_0}.R^{[a,b]}_{\bowtie r}$. The latter two formulas are defined as the difference of counterfactual probabilities (or counterfactual cumulative rewards) between interventions $I_1$ and $I_0$, in line with how causal (treatment) effects are normally estimated (see Section~\ref{sec:scms}). 

\mk{In principle, we can consider any kind of intervention $I$ over the SCM encoding of an MDP, provided that $I$ doesn't change the state space $\states$. Arguably, the most relevant case is when $I$ affects the MDP policy $\pi$. For instance, in some applications, we might want to replace $\pi$ with a more conservative or aggressive policy. Hence, in the following, unless otherwise stated, we assume interventions of the form $I=\{(\pi\gets \pi')\}$ for some $\pi'$. In the following, we discuss the the syntax and semantics of PCFTL.}

\noindent The syntax of PCFTL -- with the new terms highlighted -- is as follows:
\begin{align*}
    \Phi ::= & \top \mid \mk{\rho} \mid \neg \Phi \mid \Phi \wedge \Phi 
    \mid \fcolorbox{highlightcol1}{highlightcol}{$I_{@ t}.P_{\bowtie p}(\phi)$} \mid \fcolorbox{highlightcol1}{highlightcol}{$I_{@ t}.R^{[a,b]}_{\bowtie r}$} \mid \fcolorbox{highlightcol1}{highlightcol}{$\Delta_{@t}^{I_1,I_0}.P_{\bowtie p'}(\phi)$} \mid \fcolorbox{highlightcol1}{highlightcol}{$\Delta_{@t}^{I_1,I_0}.R^{[a,b]}_{\bowtie r}$}\\
    \phi ::= & \Phi \mid \neg \phi \mid \phi \wedge \phi \mid \phi \until{[a,b]} \phi    
\end{align*}
where $I, I_0, I_1$ are (possibly empty) interventions, $t \in \mathbb{Z}$, 
$\mk{\rho} \in \mathit{AP}$, $p\in[0,1]$, $r \in \mathbb{R}$, $p'\in [-1,1]$, $\bowtie \in \{<,\leq,\geq,>\}$, and $[a,b]$ is an interval in $\mathbb{Z}^{\geq 0}$. 

\begin{definition}[Semantics of PCFTL]\label{def:pcftl_semantics}
Given a PCFTL formula $\Phi$, an SCM $\gumbelscm{\mdp}$ for an MDP $\mdp$, and a path $\tau$ of $\gumbelscm{\mdp}$, then the \emph{PCFTL satisfaction relation} $\models$ is defined as per Definition~\ref{def:ptl_sec} and by the following rules:
\begin{align}
       (\gumbelscm{\mdp},\tau) & \models  I_{@ t}.P_{\bowtie p}(\phi) &\text{if }&
    {I}_{@t}.P_{=?}(\phi)(\gumbelscm{\mdp},\tau) \bowtie p, \label{eq:new_op_p}\\
    (\gumbelscm{\mdp},\tau) & \models  I_{@ t}.R^{[a,b]}_{\bowtie r} &\text{if }&
    {I}_{@ t}.R^{[a,b]}_{=?}(\gumbelscm{\mdp},\tau) \bowtie r, \label{eq:new_op_r}\\
    (\gumbelscm{\mdp},\tau)  & \models  \Delta_{@t}^{I_1,I_0}.P_{\bowtie p'}(\phi) & \text{if } & 
    ( {I_1}_{@t}.P_{=?}(\phi)(\gumbelscm{\mdp},\tau) - {I_0}_{@t}.P_{=?}(\phi)(\gumbelscm{\mdp},\tau) ) \bowtie p' \label{eq:effect_p}\\
    (\gumbelscm{\mdp},\tau)  &\models  \Delta_{@t}^{I_1,I_0}.R^{[a,b]}_{\bowtie r} & \text{if } & 
    ( {I_1}_{@ t}.R^{[a,b]}_{=?}(\gumbelscm{\mdp},\tau) - {I_0}_{@ t}.R^{[a,b]}_{=?}(\gumbelscm{\mdp},\tau) ) \bowtie r \label{eq:effect_r}
\end{align}
where ${I}_{@t}.P_{=?}(\phi)(\gumbelscm{\mdp},\tau)$ and ${I}_{@ t}.R^{[a,b]}_{=?}(\gumbelscm{\mdp},\tau)$ are the \emph{quantitative counterfactual operators} defined as
\begin{align}
       {I}_{@t}.P_{=?}(\phi)(\gumbelscm{\mdp},\tau) = &
    P_{\mathcal{M}'}(\{\tau' \in \textit{Paths}_{\mathcal{M}'}(s) \mid (\mathcal{M}', \tau',1) \models \phi \}) \label{eq:quant_cf_p}\\
           {I}_{@ t}.R^{[a,b]}_{=?}(\gumbelscm{\mdp},\tau) = &
    \sum_{\tau' \in \textit{Paths}_{\mathcal{M}'}(s)} \left(P_{\mathcal{M}'}(\tau')\cdot \sum_{i=a}^b R(\tau'[i])\right) \label{eq:quant_cf_r}
\end{align}
where $s=\tau[-t]$ is the state of $\tau$ at the time of intervention for $-|\tau|< t < |\tau|$, and $\mathcal{M}'=\gumbelscm{\mdp(s)}(\tau)[I]$ is the \emph{counterfactual model} starting at state $s$ given that we observed path $\tau$ and applied intervention $I$ from $s$ onward. In particular, $\mathcal{M}'$ is the SCM initialized at $S_1=s=\tau[-t]$ obtained by applying $I$ and the set $\mathbf{G}'=(G'_{s,i})_{s\in \states}^{i=1,\ldots, T-|\tau|+t}$ of \emph{posterior Gumbel variables}, where $G'_{s,i} \sim P_{\gumbelscm{\mdp}}(G_{s,i+|\tau|-t-1}\mid \tau)$ for $i=1,\ldots,t$, and $G'_{s,i} \sim \mathrm{Gumbel}$ for $i=t+1,\ldots,T-|\tau|+t$. That is, for (counterfactual) states falling within the observed path $\tau$, we use the posterior Gumbel variables inferred from $\tau$.
\end{definition}
Note that when the offset $t$ is non-negative, then the intervention state $\tau[-t]$ is the $t$-th state before the end of $\tau$, i.e., $t$ steps in the past. When $t<0$, $\tau[-t]$ is the $|t|$-th state from the start of the path.  Also, we have that $-|\tau|< t < |\tau|$, but since the length of $\tau$ is not known a priori, $t$ is clipped to $-|\tau|+1$ from below and $|\tau|-1$ from above, if needed.

\subsubsection{A generalized counterfactual operator.}
\begin{figure}
    \centering
    \subfloat[$I=\emptyset$ and $t=0$.\label{fig:trajectories_a}]{\includegraphics[width=.32\textwidth]{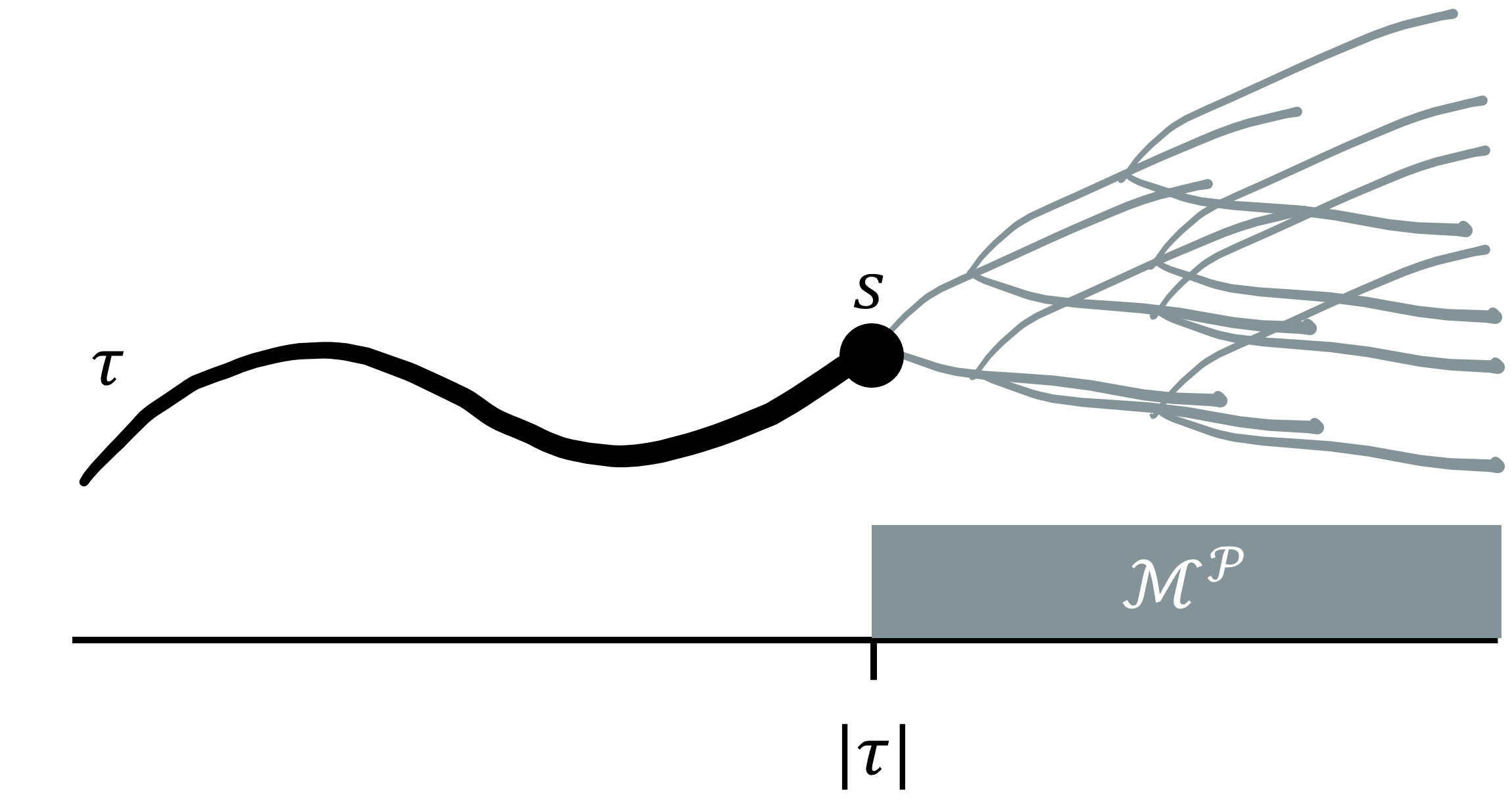}}\hfill
    \subfloat[$I\neq\emptyset$ and $t=0$.\label{fig:trajectories_b}]{\includegraphics[width=.32\textwidth]{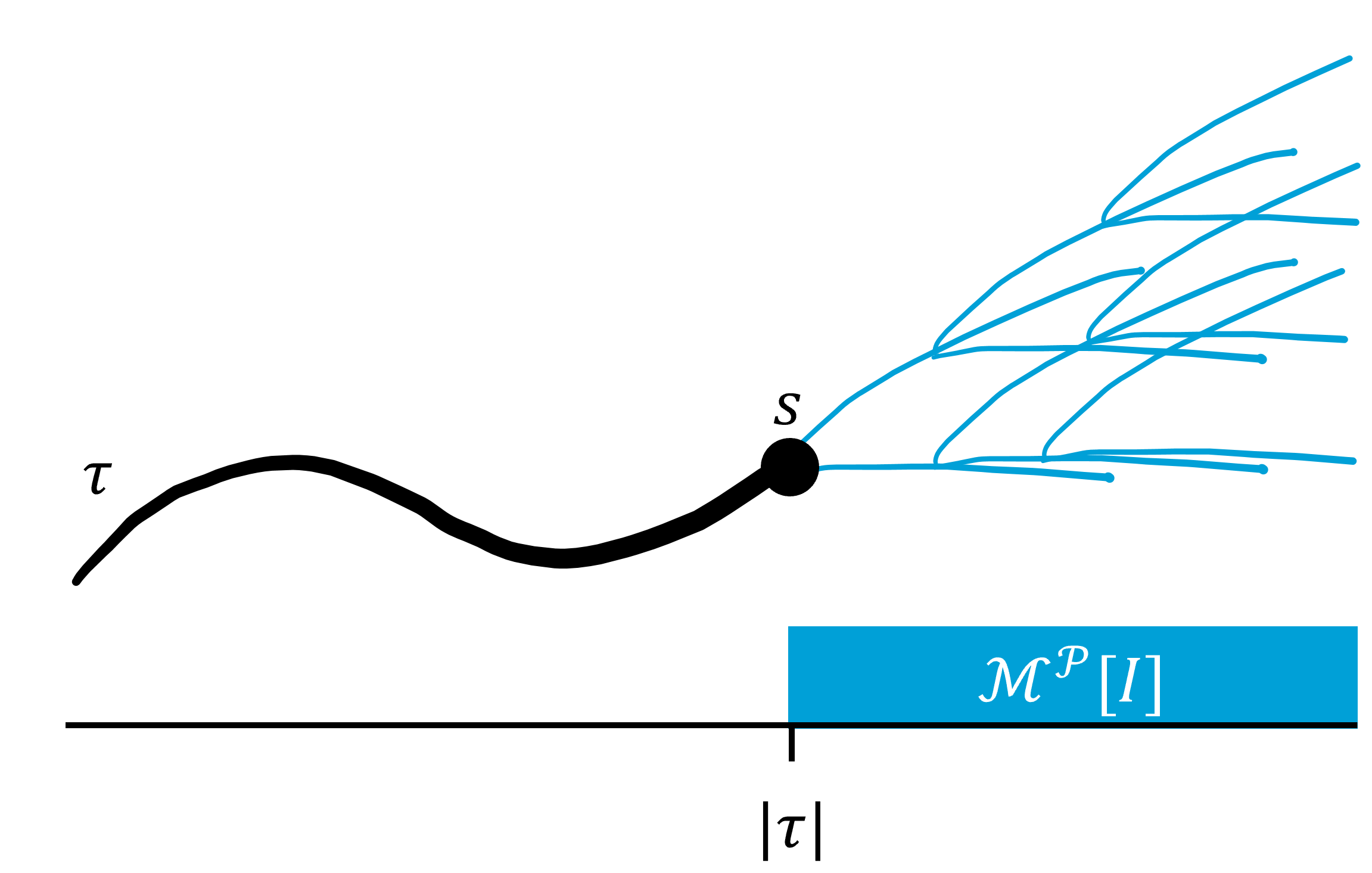}}\hfill%
    \subfloat[$I\neq\emptyset$, $t>0$
    \label{fig:trajectories_d}]{\includegraphics[width=.32\textwidth]{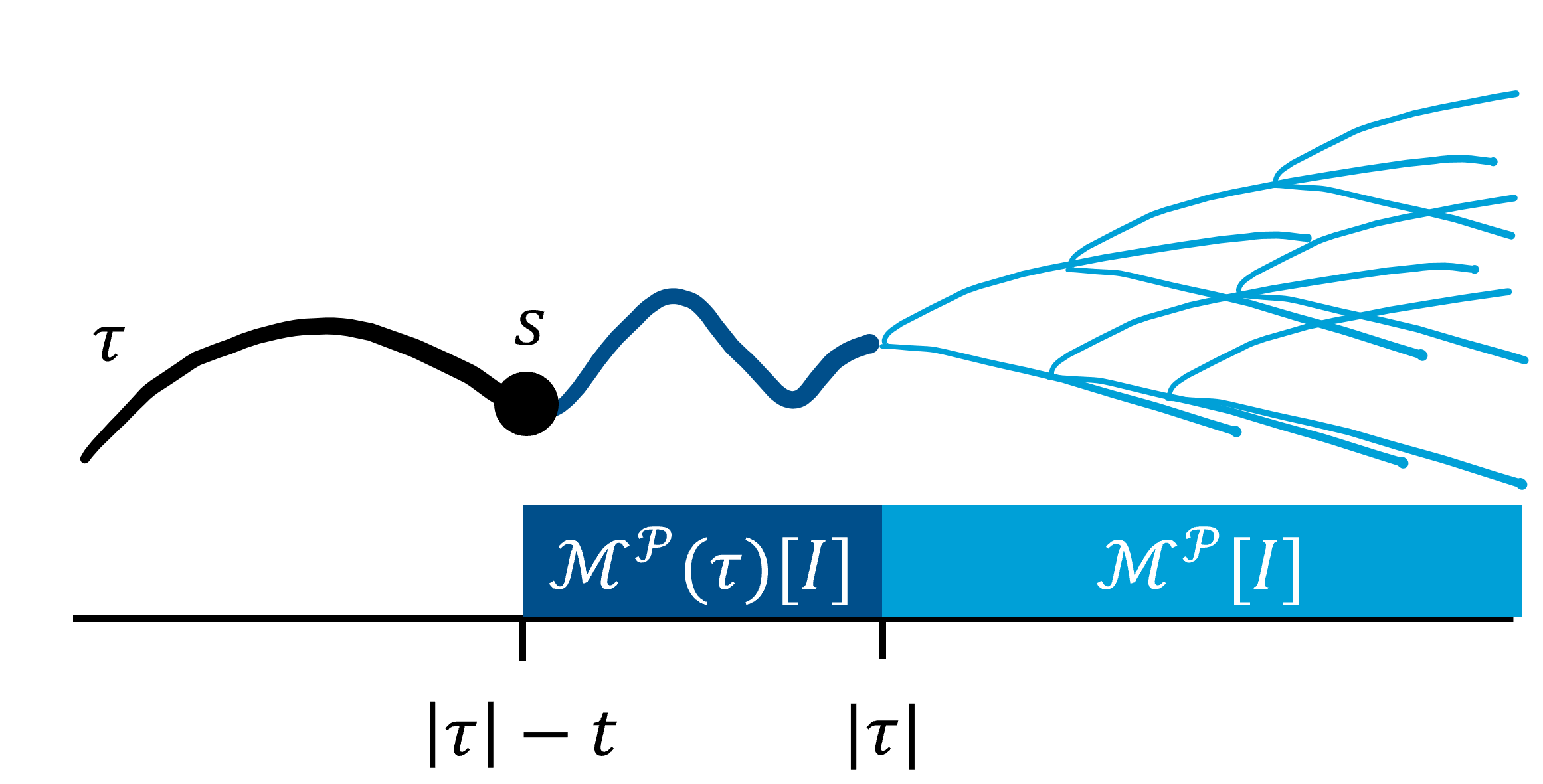}}
    \caption{\label{fig:trajectories}
    Four possible scenarios for the MDP's dynamics with
    $I_{@ t}.P_{\bowtie p}(\phi)$. The observed path $\tau$ is in black. The counterfactual path in dark blue (in general we have a posterior distribution of counterfactual paths, but here we show only one for simplicity). The set of path extensions under the nominal policy in gray, those under intervention $I$ in light blue. The horizontal axis represents time (or path positions), the vertical axis the MDP state (here is continuous and one-dimensional for illustration purposes only).}
\end{figure}

We now discuss the operator $I_{@ t}.P_{\bowtie p}(\phi)$ (a similar reasoning holds for $I_{@ t}.R^{[a,b]}_{\bowtie r}$). 
The introduced operator is general enough to capture post-interventional probabilities, that is, the probability of a path formula $\phi$ after we apply intervention $I$ at the current state. This is obtained by setting $t=0$. In this case, no counterfactuals need to be inferred because, trivially, we don't have any observed MDP states beyond the last state of $\tau$ (see also Remark~\ref{rem:abduction}). Hence, for $t=0$, we have $\gumbelscm{\mdp(s)}(\tau)[I]=\gumbelscm{\mdp(s)}[I]$, as illustrated in Fig.~\ref{fig:trajectories_b}. Also, our operator subsumes \mk{$\text{BPCTL}^{\star}$}'s probabilistic formula (which is indeed omitted in PCFTL): when $t=0$ and $I=\emptyset$, we have that $\gumbelscm{\mdp(s)}(\tau)[I]=\gumbelscm{\mdp(s)}$ (see Fig.~\ref{fig:trajectories_a}), and so, $\emptyset_{@ 0}.P_{\bowtie p}(\phi)=P_{\bowtie p}(\phi)$. 

The operator can also express the usual notion of counterfactual, which answers the question: given that we observed $\tau$, what would have been our outcome if we had applied a particular intervention $I$ in the past? This is obtained by selecting an offset $t\neq 0$.
A common choice is to apply $I$ at the beginning of $\tau$ ($t=-1$) but other options are possible, e.g., intervening before some violation has happened in $\tau$. Note that when $t\neq 0$ and $I$ is empty, then $\tau$ is the only possible path of $\gumbelscm{\mdp(s)}(\tau)[I]$ (modulo an index shift), 
see also Remark~\ref{rem:abduction}. In particular, one can see that $\emptyset_{@ t}.P_{\bowtie p}(\phi)$ is equivalent to $P_{\bowtie p}(\mathcal{P}_{[t,t]}\phi)$ where $\mathcal{P}_{[t,t]}$ is the past-equivalent operator of $\finally{[t,t]}$.

An important feature of our operator, when $t\neq 0$, is that it allows to simultaneously reason about the counterfactual past and the future evolution of the MDP under $I$ from the counterfactual present. That is, depending on the bounds in the temporal operators of the path formula $\phi$, evaluating $\phi$ might require paths that extend beyond $\tau$. Hence, up to the length of $\tau$, realizations of $\gumbelscm{\mdp}(\tau)[I]$ correspond to counterfactual paths, as described above; beyond the length of $\tau$, realizations follow the post-interventional model $\gumbelscm{\mdp}[I]$ because there are no observations to condition on (and hence $\gumbelscm{\mdp}(\tau)[I]=\gumbelscm{\mdp}[I]$). This behaviour is illustrated in Fig.~\ref{fig:trajectories_d}. Therefore, we generalize the usual approach to counterfactual inference of MDPs~\cite{oberst2019counterfactual,tsirtsis2021counterfactual} which doesn't consider the future behaviour of the MDP beyond the counterfactual path. We show why this matters with the example below. 

\begin{example}\label{ex:cf_operator} Consider a Gumbel-Max SCM $\gumbelscm{\mdp}$ for an MDP $\mdp$ and an obstacle avoidance property $\phi_H = \globally{[0,H]} \neg \mathrm{obstacle}$ for some horizon $H>0$. Let $\tau$ be the observed path and $\tau_I \sim P_{\gumbelscm{\mdp(\tau[1])}(\tau)[I]}(S_1,A_1,\ldots,S_{|\tau|},A_{|\tau|})$ be the corresponding counterfactual path under some intervention $I$ applied at the beginning of $\tau$. Suppose that no obstacle is hit in $\tau$ or $\tau_I$. So, in usual counterfactual analysis, one can conclude that the nominal policy and the intervention policy are equivalent relative to property $\phi_H$. However, if the safety property bound $H$ extends beyond the length $|\tau|$ of the observed path, then it is necessary to reason about the evolution of the MDP going forward: in one case, starting from the last state of $\tau$ and under the nominal policy; in the other, from the last state of the counterfactual path $\tau_I$ and under $I$'s policy. At this point, it is entirely possible that $\emptyset_{@(-1)}.P_{=?}(\phi_H)(\gumbelscm{\mdp},\tau)<I_{@(-1)}.P_{=?}(\phi_H)(\gumbelscm{\mdp},\tau)$, i.e., going forward from the counterfactual present yields a higher probability of obstacle avoidance than remaining with the nominal policy. Thus, limiting the analysis to the counterfactual past only, as done in previous work~\cite{oberst2019counterfactual,tsirtsis2021counterfactual}, would lead to the wrong conclusion that the two policies are equivalent safety-wise.
\end{example}

\subsubsection{Encoding Treatment Effects.} We explain how the introduced causal effect operators $\Delta_{@t}^{I_1,I_0}.P_{\bowtie p'}(\phi)$ and $\Delta_{@t}^{I_1,I_0}.R^{[a,b]}_{\bowtie r}$ can be used to express the traditional (C)ATE and ITE estimators (defined in Section~\ref{sec:scms}). 
We saw that CATE is the difference of post-interventional probabilities, conditioned on a particular value $V=v$ of some variable $V$. In reinforcement learning with MDPs, one sensible choice is to condition on the first state of the post-interventional path~\cite{oberst2019counterfactual}. Therefore, for the same argument made above about defining post-interventional probabilities with $I_{@ 0}.P_{\bowtie p}(\phi)$ formulas, we can express this notion of CATE in PCFTL with the formula $\Delta_{@0}^{I_1,I_0}.P_{\bowtie p}(\phi)$. The latter indeed is the effect in the probability of $\phi$ between interventions $I_1$ and $I_0$, conditioned on paths starting with $\tau[0]$ (the last state of $\tau$), where $\tau$ is the observed path prefix. 

ATE, the unconditional version of CATE, cannot be directly expressed in PCFTL because the semantics is defined over a non-empty path $\tau$, and hence, probabilities are implicitly conditional on state $\tau[-t]$. An equivalent of ATE can be defined, given some distribution $P$ of MDP states and $S\sim P(S)$, as the expectation w.r.t.\ $S$ of the above CATE definition, conditional on $S$: 
$\mathbb{E}_{S\sim P(S)}\left[\Delta^{I_1,I_0}_{@0}.P_{=?}(\phi)(\gumbelscm{\mdp},(S))\right]$.

Finally, akin to how ${I}_{@t}.P_{\bowtie p}(\phi)$ with $t\neq 0$ provides a generalized notion of counterfactual probability (see previous subsection), the operator $\Delta_{@t}^{I_1,I_0}.P_{\bowtie p}(\phi)$ with $t\neq 0$ provides a generalized notion of ITE, because, like ITE, our operator is defined as the difference of counterfactual probabilities  ${I_1}_{@t}.P_{=?}(\phi)$ and ${I_0}_{@t}.P_{=?}(\phi)$.

\subsubsection{Example Properties.}
Let $\phi$ be a path formula describing some requirement of interest. Let $\pi$ be the nominal policy of $\gumbelscm{\mdp}$ and $I'=\{(\pi\gets \pi')\}$ be the intervention where we apply some policy $\pi'$. Some useful properties that can be specified in PCFTL are: 
\begin{itemize}
    \item \emph{Is $\pi'$ safer than $\pi$ moving forward from the current state?} This is a CATE-like query, which we can express as $\Delta^{I',\emptyset}_{@ 0}. P_{>0}(\globally{[a,b]}\phi)$ for some bounds $a$ and $b$.
    \item {A regret monitor.} Suppose we require a safety probability of at least $p$. Then we can ask: \emph{had we deployed $\pi'$ $t$ steps in the past, would have we observed a safety probability of at least $p$ when $\pi$ fails to achieve so?} This is expressed as: $$\emptyset_{@ t}. P_{<p}(\mathcal{G}_{[a,b]} \phi) \rightarrow I'_{@ t}. P_{\geq p}(\mathcal{G}_{[a,b]} \phi).$$
    \item A nested formula. \textit{What would have been the probability, had we applied $\pi'$ $t$ steps in the past, of observing a violation at a given time point $t'$, and after this point, of a different policy $\pi''$ yielding a better recovery probability than $\pi'$? }
    $$I'_{@ t}. P_{=?}(\mathcal{F}_{[t',t']} (\neg \phi \wedge \Delta^{I'',\emptyset}_{@ 0}. P_{>0} \mathcal{F}_{[1,H]}\phi)) \text{, where $I''=\{(\pi\gets \pi'')\}$ and $H\geq 1$}.$$ 
\end{itemize}

%% file: procedures.tex
\subsection{Decision procedures}\label{sec:decision}

We take a statistical model checking approach~\cite{younes2006statistical,legay2010statistical} to decide satisfaction of our properties, i.e., based on Monte-Carlo sampling of the Gumbel-Max SCM model. 
We leave the investigation of numerical-symbolic decision procedures for future work. 

We restrict to non-nested properties, i.e., those where path sub-formulas $\phi$ don't contain in turn counterfactual operators, even though we allow for arbitrary nesting of temporal operators in $\phi$. The complication with nested formulas is that the satisfaction of $\phi$ cannot be determined by a single execution and is subject to statistical errors. There are ways to compute \textit{a-priori} error bounds in the nested case as well~\cite{younes2006statistical,legay2010statistical}, but the approach remains inefficient because it requires a number of realizations exponential in the depth of the nested operator.
For space reasons, 
we provide a complete description of the statistical model checking procedures in Appendix~\ref{app:decision}.

%% file: experiments.tex
\section{Experimental Evaluation}\label{sec:results}
We provide two sets of results. In the first one, we consider a simple grid-world model and a reach-avoid specification. We use this case study to provide a detailed analysis of interventional and counterfactual probabilities, their variability, and the accuracy of counterfactual inference. In the second set of results, we apply PCFTL to a benchmark of more complex 2D grid-world environments selected from the MiniGrid library~\cite{minigrid}.
\input{reach-avoid}
\input{minigrid}

%% file: reach-avoid.tex
\subsubsection{Reach-Avoid Example.}
We consider a $4\times 4$ grid-world example, where 
the agent can move up, down, left, or right, one square at a time. The specification $\phi$ is one of reach-avoid: we want to reach some goal region while avoiding an unsafe region, i.e., $ \phi \equiv \neg \mathrm{unsafe} \ \until{[0,T]} \mathrm{goal}$. We choose $T=10$. 
We consider two policies, a nominal (default) policy $\pi$ and an optimal policy $\pi_o$.
The optimal policy is 
found by value iteration after assigning a reward $1$ to the goal and making the unsafe and goal states terminal. The nominal policy is defined manually to make it intentionally less safe than $\pi_o$.
The stochasticity comes from the fact that the agent, with small probability ($0.1$ in our experiments), randomly takes a different action than that of the policy. 

\begin{figure}
    \centering
    \subfloat[$\pi_o$ applied at the\\ beginning.\label{fig:from_start_optimal_vs_random}]{\includegraphics[width=0.32\textwidth, height=3cm]{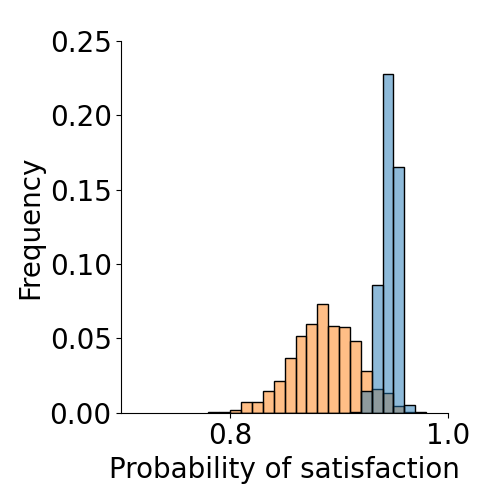}}
    \hfill
    \subfloat[$\pi_o$ applied after the\\ first step\label{fig:from_first_step_optimal_vs_random}]{\includegraphics[width=0.32\textwidth, height=3cm]{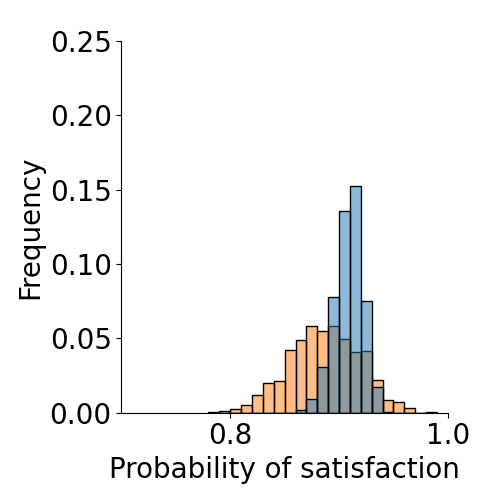}}
        \hfill
    \subfloat[$\pi_o$ applied after the first step, on  paths shorter than $T$. \label{fig:from_first_step_optimal_vs_random_intrv}]{
    \includegraphics[width=0.32\textwidth, height=3cm]{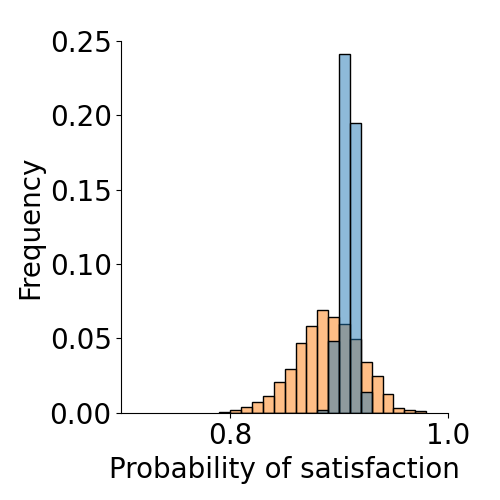}
    }
     \caption{Counterfactual probabilities under the optimal policy $\pi_o$ (blue) given that we observe MDP paths under the nominal policy $\pi$ (orange). In (a) and (b) paths have length $10$ (same as the time bound $T$ in $\phi$). In (c), we observe paths of length $2<T$, and so, applying $\pi_o$ results in paths that are part counterfactual, part post-interventional.}
    \label{fig:policies}
\end{figure}

For each experiment in this subsection, we perform $1,000$ repetitions to evaluate the variability of the probability estimates. For each repetition, we generate $100$ observed paths under the nominal policy. 
Counterfactuals are estimated using $20$ posterior Gumbel realizations. 
Probability values are computed by averaging the satisfaction value of $\phi$ over all paths within each repetition. We choose the optimal policy as the interventional/counterfactual one, by defining $I=\{\pi \gets \pi_o\}$. 
We evaluate the performance of the optimal policy in a counterfactual setting. In particular, we compare the probability $P_{=?}(\phi)$ under the nominal MDP against the average counterfactual probability $I_{@ t}.P_{=?}(\phi)$ for some $t$ (where the average is w.r.t.\ the set of nominal paths used for $P_{=?}(\phi)$). We apply $I$ at the beginning of the path ($t=-1$, see Fig.~\ref{fig:from_start_optimal_vs_random}) and after the first step ($t=-2$, see Fig.~\ref{fig:from_first_step_optimal_vs_random}). 
Since $\pi_o$ (blue histograms in Fig.~\ref{fig:policies}) is safer than $\pi$ (orange), the distribution of counterfactual probabilities clearly dominates that under nominal settings. 
See Fig.~\ref{fig:from_start_optimal_vs_random}. For the same reason, delaying the intervention of one step leads to more unsafe trajectories (the blue histogram in Fig.~\ref{fig:from_first_step_optimal_vs_random} is indeed shifted to the left compared to that in Fig.~\ref{fig:from_start_optimal_vs_random}). 

\review{In Fig.~\ref{fig:from_first_step_optimal_vs_random_intrv} we provide results of a generalized counterfactual query, that is, involving both the counterfactual past and the subsequent future evolution of the system.}  
To do so, we draw paths  $\tau$ under $\pi$ of length $2$ and apply $I$ after the first time step. This results in paths that are counterfactual in the first part (because we apply $I$ in the past, conditioned by $\tau$) and post-interventional 
in the second part (because to evaluate $\phi$, we need paths longer than the observed $\tau$). 

%% file: minigrid.tex
\subsubsection{MiniGrid Benchmark.}
 MiniGrid~\cite{minigrid} is a collection of 2D grid-world environments with goal-oriented tasks designed for developing reinforcement learning algorithms.
 Each cell in this grid world is encoded as a three-dimensional tuple (object, color,  state). There are $8$ different objects, $6$ colors and $3$ states: open, close and locked. There are $7$ actions that the agent can take which are turn left, turn right, move forward, pick up, drop, toggle and done. 
 We consider four of these environments: \textit{Empty}, \textit{DoorKey}, \textit{GoToDoor} and \textit{Fetch}. 

\textit{Empty} is the simplest environment, where the agent simply navigate the grid to reach some goal. This corresponds to the specification $\finally{[0,T]} \mathrm{goal}$, where we choose $T=50$. In the \textit{DoorKey} environment, a key and a door exist on the grid. The agent must first find the key, unlock the door, and reach the goal, expressed as
$\phi \equiv \finally{[0,T_k]}(\mathrm{key} \wedge \finally{[0,T_d]}(\mathrm{door} \wedge \finally{[0,T_g]} \mathrm{goal}))$, with $T_k+T_d+T_g=T$. 
This task requires the agent to learn basic navigation skills and non-trivial sequential plans. In \textit{GoToDoor}, we have four doors with different colors, and the agent is tasked to reach the door of some given color: 
$\phi \equiv \finally{[0,T]}\mathrm{door}$. 
The door is always unlocked, making it a simpler task than \textit{DoorKey}. 
 In \textit{Fetch}, the grid contains multiple objects with assorted colors which the agent must pick up and bring to the goal: 
 $\phi \equiv \finally{[0,T_o]}(\mathrm{object} \wedge \mathcal{X}(\mathrm{carrying}\ \mathcal{U}_{[0,T_g]} \mathrm{goal}))$, with $T_o+T_g+1=T$.
This task requires the agent to learn to manipulate objects and navigate the grid. 

For each environment, we train two convolutional neural network policies using the Proximal Policy Optimization (PPO)~\cite{schulman2017proximal} algorithm. 
For the nominal policy $\pi$, this time we use an optimal policy, trained using $10$ million time steps. The interventional/counterfactual policy is intentionally undertrained, using only $200$ time steps. 

Experimental results are presented in Table~\ref{tab:results_of_experiments}. We examine probabilities under the nominal/optimal policy (2nd column) using the formula $P_{\geq 0.9}(\phi)$, counterfactual probabilities with the undertrained policy  (3rd column) using ${I}_{@t}.P_{\geq 0.9}(\phi)$, and determine the causal effect between the two (4th column) using $\Delta_{@t}^{I,\emptyset}.P_{>0}(\phi)$. For every environment and PCFTL formula, we carry out two set of experiments using two different intervention points, at the start of the trajectory ($t=T$), and 10 steps into the trajectory ($t=T-10$).

Verification results are computed using statistical model checking (see Appendix~\ref{app:decision} for details on the decision procedures). Results indicate that the system does not satisfy the property when using an undertrained policy, while the optimal policy is always successful. This performance gap can also be seen in the causal effect column. We observe that the verification procedure is very efficient (requiring at most $125$ realizations), and that the number of realizations necessary to obtain a positive answer for the nominal policy is higher than those for a negative answer for the interventional policy. The reason is that we set a high probability threshold, $p\ge 0.9$, and so, even with a well-trained policy, we require a fair amount of evidence to conclude that the property is satisfied. Similarly, the interventional policy performs enough badly to require much fewer points for concluding that the property is violated.

\begin{table}
    \caption{PCFTL verification of the MiniGrid benchmark, with 6x6 grids. For each environment, we have two sets of experiments: with intervention at the start of the path ($t=T$) and $10$ steps after the start ($t=T-10$), with $T=50$ being the length of the path. The parameters of the statistical model checking algorithm (see Appendix~\ref{app:decision}) are $\delta=0.02$, and $\alpha= 0.05$ and $\beta=0.2$ for $P$ and ${I}_{@t}.P$ properties, and $\alpha= 0.01$ and $\beta=0.2$ for $\Delta_{@t}^{I,\emptyset}.P$. In parentheses are the number of Monte-Carlo realizations required by the verification algorithm to reach a verdict.}
    \label{tab:results_of_experiments}
    \centering
    \begin{tabular}{c|c|c|c|c}
       Environment & $t$  &  $P_{\geq 0.9}(\phi)$ & ${I}_{@t}.P_{\geq 0.9}(\phi)$ & $\Delta_{@t}^{I,\emptyset}.P_{>0}(\phi)$\\
       \hline
       DoorKey6x6&$T$&  \multirow{2}{*}{$\top$(125)} & $\bot$(25) & $\bot$(50) \\
DoorKey6x6&$T-10$&   & $\bot$(25) & $\bot$(50) \\
\hline
Empty6x6&$T$&  \multirow{2}{*}{$\top$(75)} & $\bot$(25) & $\bot$(50) \\
Empty6x6&$T-10$&  & $\bot$(25) & $\bot$(75) \\
\hline
Fetch6x6&$T$&  \multirow{2}{*}{$\top$(75)} & $\bot$(25) & $\bot$(75) \\
Fetch6x6&$T-10$&  & $\bot$(50) & $\bot$(75) \\
\hline
GoToDoor6x6&$T$&  \multirow{2}{*}{$\top$(125)} & $\bot$(25) & $\bot$(50) \\
GoToDoor6x6&$T-10$& & $\bot$(25) & $\bot$(100)
    \end{tabular}
\end{table}

%% file: related.tex
\section{Related Work}\label{sec:related}

\paragraph{Causality and Verification.} Concepts of causality have been investigated in formal verification for years~\cite{baier2021verification}. Two main classes of approaches exist, respectively based on the theory of actual causality ~\cite{halpern2016actual,halpern2020causes} and on probabilistic causation. 
Given an SCM $\mathcal{M}$ and a context $\mathbf{u}$, an actual cause is, informally, the smallest set of SCM variables that, if forced with a different value, lead to a different (counterfactual) outcome for some target variable $Y$. 
This notion has been adapted in~\cite{beer2012explaining} to find so-called root causes in LTL counterexample traces and in~\cite{gossler2014blaming} to identify the components of a timed-automata network responsible for a given failure trace. 
Probabilistic causation methods like~\cite{baier2021probabilistic,baier2022probability,kleinberg2009temporal,kleinberg2011logic} build on the probability-raising (PR) principle by which the probability of an effect $E$ is higher after observing a cause $C$ than if the cause had not happened. More precisely, these works consider Markov models and express $E$ and $C$ as sets of states or PCTL state formulas. 

Our work is complementary to these methods in that our concern is not to identify causes given some observations, but to reason about the probability of a temporal logic specification in interventional and counterfactual settings.
Methods based on  actual causality similarly rely on counterfactuals but consider only non-probabilistic models. Methods based on the PR principle support probabilistic models but do not allow for model manipulations, and so they cannot reason about counterfactual outcomes.

\review{\paragraph{Probabilistic hyperproperties.} 
Probabilistic hyper-properties (PHPs) for MDPs have been recently introduced in~\cite{dimitrova2020probabilistic,abraham2020probabilistic} to support quantification over MDP schedulers (i.e., policies). One can see that PHPs for MDPs are strictly more expressive than the fragment of PCFTL without counterfactuals (i.e., where interventions can be applied only at $t=0$). For instance, the PCFTL causal-effect formula $\Delta^{I_1,I_0}_{@0}.P_{\bowtie p}(\phi)$ can be expressed as the PHP $\exists \sigma_1 \exists \sigma_0 . P(\phi_{\sigma_1 }) - P(\phi_{\sigma_0 }) \bowtie p$ (using the syntax of~\cite{dimitrova2020probabilistic}) where the domains of schedulers $\sigma_0$ and $\sigma_1$ are singletons (and chosen to be consistent with $I_0$ and $I_1$, respectively). However, PHPs do not support counterfactuals, which is arguably the main strength of our method.
}

\paragraph{Causality in Reinforcement Learning.} There is a growing interest in applying causal inference in RL, for instance, to evaluate counterfactual policies from observational data~\cite{oberst2019counterfactual}, provide counterfactual explanations~\cite{tsirtsis2021counterfactual} (i.e., the minimum number of policy actions to change in order to attain a better outcome), produce counterfactual data to enhance training of RL policies~\cite{forney2017counterfactual,buesing2018woulda}, or estimate causal effects in presence of confounding factors~\cite{lu2018deconfounding}. These works are very relevant yet they consider different problems from ours.
That said, PCFTL builds on~\cite{oberst2019counterfactual} where the authors introduce Gumbel-Max SCMs and prove the counterfactual stability property. 

%% file: conclusion.tex
\section{Conclusion}\label{sec:conclusion}
We have presented the probabilistic temporal logic PCFTL, the first of its kind to enable causal reasoning about interventions, counterfactuals and causal effects in Markov Decision Processes. From a syntactic viewpoint, this is achieved by introducing an operator that subsumes interventions, counterfactuals and the traditional probabilistic operator. We defined the semantics of PCFTL in terms of Gumbel-Max structural causal models, which provide an encoding of discrete-state MDPs amenable to counterfactual reasoning. We performed a set of experiments on a benchmark of grid-world models, demonstrating the usefulness of the approach (being applicable to deep reinforcement learning policies as well) and the accuracy of counterfactual inference. 
We envision several future  directions for this work, including investigating symbolic (as opposed to statistical) model checking algorithms, partial observability, and extending the logic to different classes of systems, like cyber-physical and data-driven systems.

%% file: appendix.tex
\newpage\appendix
\section{Counterfactual Stability}\label{app:cf_stab}
We describe a desirable property for an SCM, called \textit{counterfactual stability}~\cite{oberst2019counterfactual}. This property pertains to discrete SCM variables, whose distribution is thus categorical. Let $Y$ be the (discrete) variable of interest. Let $\mathbf{p}$ be the vector of probabilities for $P_{\scm[I]}(Y)$, i.e., the  (categorical) distribution of $Y$ under some intervention $I$,
and let $\mathbf{p}'$ denote the vector of probabilities for $P_{\scm[I']}(Y)$. Let $P_{\scm[I](Y=i)[I']}(Y)$ be the counterfactual distribution of $Y$ under $I'$ given that we observed $Y=i$ under $I$.

\begin{definition}[Counterfactual stability~\cite{oberst2019counterfactual}]
An SCM $\scm$ satisfies \emph{counterfactual stability} relative to a categorical variable $Y$ of $\scm$ if the following holds. If we observe $Y=i$ in a realization of $P_{\scm[I]}(Y)$, then for all $j\neq i$, if $\dfrac{p_i'}{p_i}\geq \dfrac{p_j'}{p_j}$ then $P_{\scm[I](Y=i)[I']}(Y=j)=0$. That is, if we observed $Y=i$ under intervention $I$ then the counterfactual value of $Y$ under $I'$ cannot be equal to $j\neq i$ unless the multiplicative change in $p_i$ is less than the multiplicative change in $p_j$. 
\end{definition}
\begin{corollary}[Stability and invariance of counterfactuals~\cite{oberst2019counterfactual}]
Let $\scm$ be an SCM that satisfies \emph{counterfactual stability}.  If we observe $Y=i$ under $\scm[I]$ and $\dfrac{p_i'}{p_i}\geq \dfrac{p_j'}{p_j}$ holds for all $j\neq i$, then $P_{\scm[I](Y=i)[I']}(Y=i)=1$.
\end{corollary}
Intuitively, the above definition and corollary tell us that, in a counterfactual scenario, we would observe the same outcome $Y=i$ unless the intervention increases the relative likelihood of an alternative outcome $Y=j$, that is, unless $\dfrac{p_j'}{p_j}>\dfrac{p_i'}{p_i}$ holds for some $j$.

%% file: procedures_long.tex
\section{Decision procedures}\label{app:decision}

Before discussing the decision procedures, below we reformulate counterfactuals and causal effects in a way that facilitates statistical model checking. 

\subsubsection{Computation of Counterfactuals and Causal Effects.} 
We can express the counterfactual probability of~\eqref{eq:quant_cf_p} as the expectation of a function $f(\mathbf{G}')$ of the posterior Gumbel $\mathbf{G}'$, as follows:
\begin{equation}\label{eq:alt_cf_p}
    I_{@ t}.P_{=?}(\phi)(\gumbelscm{\mdp},\tau) = \mathbb{E}_{\mathbf{G}'}[f(\mathbf{G}')], \text{ with } f(\mathbf{G}')=\mathbf{1}((\gumbelscm', \tau'(\mathbf{G}'), 1) \models \phi),
\end{equation}
where $\mathbf{1}$ is the indicator function, $\mathcal{M}'=\gumbelscm{\mdp(s)}(\tau)[I]$ and $\mathbf{G}'$ are, respectively, the counterfactual model and posterior Gumbel (defined in Definition~\ref{def:pcftl_semantics}), and $\tau'(\mathbf{G}')$ is the path of $\mathcal{M}'$ uniquely determined by $\mathbf{G}'$\footnote{This decomposition is analogous to how we obtain the distribution $P_{\scm}$ of an SCM $\scm$ as a function of the distribution $P(\exovars)$ of its exo-variables $\exovars$.}. This formulation is convenient for our statistical decision procedures, allowing us to sample realizations of the counterfactual outcome by first sampling from the distribution of $\mathbf{G}'$ and then applying $f$. The corresponding formulation for the counterfactual reward of~\eqref{eq:quant_cf_r} is readily obtained as:
\begin{equation}\label{eq:alt_cf_r}
    R^{[a,b]}_{=?}(\gumbelscm{\mdp},\tau) = \mathbb{E}_{\mathbf{G}'}[f(\mathbf{G}')], \text{ with } f(\mathbf{G}')=\sum_{i=a}^b R(\tau'(\mathbf{G}')[i]).
\end{equation}
We proceed similarly for causal effect operators, with one important difference. While Equations~\ref{eq:effect_p} and~\ref{eq:effect_r} define the causal effect as the difference of two independent probabilities (or expected rewards), we express it as the mean of paired differences between individual outcomes. This is possible because the two counterfactual models $\gumbelscm{\mdp}(\tau)[I_1]$ and $\gumbelscm{\mdp}(\tau)[I_0]$ share the same distribution of posterior Gumbel (only the intervention changes). Below we provide $f(\mathbf{G}')$ for the two causal effects operators. 
\begin{align}
    \Delta^{I_1,I_0}_{@t}.P_{=?}(\phi)(\gumbelscm{\mdp},\tau): & \ \  f(\mathbf{G}')=\mathbf{1}((\gumbelscm{}_1, \tau_1(\mathbf{G}'), 1) \models \phi) - \mathbf{1}((\gumbelscm{}_0, \tau_0(\mathbf{G}'), 1) \models \phi) \label{eq:alt_effect_p}\\
    \Delta^{I_1,I_0}_{@t}.R^{[a,b]}_{=?}(\gumbelscm{\mdp},\tau) : & \ \ f(\mathbf{G}')=\sum_{i=a}^b R(\tau_1(\mathbf{G}')[i])-R(\tau_0(\mathbf{G}')[i]),\label{eq:alt_effect_r}
\end{align}
where $\mathcal{M}_i=\gumbelscm{\mdp}(\tau)[I_i]$ is the counterfactual model for $I_i$, and $\tau_{i}(\mathbf{G}')$ is the path of $\mathcal{M}_i$ uniquely determined by $\mathbf{G}'$. The advantage of the above form using paired differences is that this yields smaller variability, and hence, a more accurate statistical estimation, than the one based on the difference of independent means.

\subsection{Qualitative Properties}
Let $p_{\phi}=I_{@t}.P_{=?}(\phi)(\gumbelscm{\mdp},\tau)$ be the true (unknown) counterfactual probability of $\phi$ for a given MDP $\mdp$ and path $\tau$.  The problem of checking whether $p_{\phi}$ is above a given threshold $\theta$, i.e., deciding property $I_{@t}.P_{\geq \theta}(\phi)$, can formulated and solved as one of hypothesis testing, where we test the hypothesis $H: p_{\phi}\geq \theta$ against $K: p_{\phi}< \theta$ using a set of observations $x_1, \ldots, x_m$ of the underlying process.  

Hypothesis testing may incur two kinds of errors: \textit{type-1 errors}, i.e., wrongly concluding that $K$ is true (when $H$ holds) and \textit{type-2 errors}, i.e., wrongly concluding that $H$ is true (when $K$ holds). We denote the probability of \textit{type-1 errors} by $\alpha$ and that of  \textit{type-2 errors} by $\beta$. The pair $\langle \alpha,\beta \rangle$ is also called the \textit{strength} of the test.

Wald's sequential probability ratio test (SPRT)~\cite{wald2004sequential} is an efficient scheme used in probabilistic model checking~\cite{younes2006statistical,younes2006numerical} to sample only the number of realizations necessary to answer the above hypothesis test with strength $\langle \alpha,\beta \rangle$. We first explain in detail the SPRT for $I_{@t}.P_{\geq \theta}(\phi)$ properties, and then briefly cover the other kinds of formulas. 

\paragraph{$I_{@t}.P_{\geq \theta}(\phi)$ properties.}
The SPRT method considers the following relaxation of the original hypotheses:
$H_0: p_{\phi}\geq \theta_0 \text{ VS } H_1: p_{\phi}\leq \theta_1, \text{ with } \theta_0=\theta+\delta \text{ and } \theta_1=\theta-\delta,$ 
where $\delta>0$ is a user-defined parameter. The interval $(\theta_1,\theta_0)$ is called \textit{indifference region}, as we are willing to accept either hypothesis when $p_{\phi} \in (\theta_1,\theta_0)$. 
This relaxation is necessary because, when testing the original hypotheses $H$ and $K$, we cannot control simultaneously both $\alpha$ and $\beta$ if the true probability $p_{\phi}$ is exactly equal to $\theta$, see~\cite{younes2006statistical,younes2006numerical}. 

In the SPRT, we collect observations iteratively. At the $m$-th iteration, we have $m$ observations $\mathbf{x}_m = (x_1,\ldots,x_m)$. In our case, these are counterfactual outcomes, i.e., realizations of the Bernoulli process $(X_1,\ldots, X_m)$ where $X_i \sim f(\mathbf{G'})=\mathbf{1}((\gumbelscm', \tau'(\mathbf{G}'), 1) \models \phi)$ (see Eq.~\ref{eq:alt_cf_p}). 
Given $\mathbf{x}_m$, we compute the following likelihood ratio (LR)
\begin{equation}\label{eq:lh_probs}
    \dfrac{f(\mathbf{x}_m \mid H_1)}{f(\mathbf{x}_m \mid H_0)}=\dfrac{\prod_{i=1}^m Pr(X_i = x_i \mid p_{\phi}=\theta_1)}{\prod_{i=1}^m Pr(X_i = x_i \mid p_{\phi}=\theta_0)}=\dfrac{\theta_1^{d_m}(1-\theta_1)^{m-d_m}}{\theta_0^{d_m}(1-\theta_0)^{m-d_m}},
\end{equation}
where $d_m=\sum_{i=1}^m x_i$ is the number of observed successes. In other words, $f(\mathbf{x}_m \mid H_i)$ is the probability of observing the sequence $\mathbf{x}_m$ if $p_{\phi}=\theta_i$ holds. At this point, the SPRT compares the LR with the constants $A=(1-\beta)/{\alpha}$ and $B={\beta}/{(1-\alpha)}$ and:
\begin{itemize}
    \item if $\dfrac{f(\mathbf{x}_m \mid H_1)}{f(\mathbf{x}_m \mid H_0)}\leq B$, we accept $H_0$, with a type-2 error probability of $\beta$;
    \item if $\dfrac{f(\mathbf{x}_m \mid H_1)}{f(\mathbf{x}_m \mid H_0)}\geq A$, we accept $H_1$, with a type-1 error probability of $\alpha$; or,
    \item we collect additional observations until one of the two above conditions hold. 
\end{itemize}
Note that this procedure requires a larger number of observations as the true $p_{\phi}$ approaches the threshold $\theta$. Nevertheless, a decision is always reached after a finite number of steps (see~\cite{younes2006statistical,younes2006numerical} for a more detailed analysis of the SPRT's stopping time). The above decision scheme is valid for other kinds of properties as well, i.e., it doesn't depend on the underlying distribution of the observations, as long as the LR is adequately defined. Hence, we won't repeat it for the cases below.

\paragraph{$I_{@ t}.R^{[a,b]}_{\geq \theta}$ formulas.} The SPRT can be also applied to variables other than Bernoulli, as are those entailed by reward-based properties. The corresponding test is an application of the SPRT to T-distributed observations~\cite{schnuerch2020controlling}. 
Let $\mu$ be the true (unknown) average cumulative reward, i.e., $\mu=I_{@ t}.R^{[a,b]}_{=?}$.  
Here, we sample observations from the $f(\mathbf{G}')$ of Eq.~\ref{eq:alt_cf_r}, for which we have that $\mu=\mathbb{E}[f(\mathbf{G}')]$. 

We consider the hypotheses:  
$H_0: \mu\geq \theta_0 \text{ VS } H_1: \mu \leq \theta_1, \text{ with } \theta_0 = \theta + \delta\cdot \sigma \text{ and } \theta_1 = \theta - \delta\cdot \sigma,$ 
where $\sigma$ is the (unknown) standard deviation of $f(\mathbf{G}')$, and $\delta>0$ is the indifference parameter: the indifference region spans $2\cdot \delta$ standard deviations around $\theta$.  

The definition of the LR follows the intuition that if $H_0$ holds and in particular, $\mu=\theta_0$, then the variable $T_m = (\bar{X}_m-\theta)/{S_m}$ follows a non-central T distribution with non-centrality parameter $\delta\cdot\sqrt{m}$ and $m-1$ degrees of freedom, where $\bar{X}_m=\frac{1}{m}\sum_{i=1}^m X_i$ is the sample mean and $S_m = \frac{1}{\sqrt{m}}\sqrt{\frac{1}{m-1} \sum_{i=1}^m (X_i - \bar{X}_m)^2}$ is the standard error of $\bar{X}_m$~\cite{schnuerch2020controlling}. The same reasoning holds for $H_1$, but after adjusting the sign of $T_m$. 
Hence, the LR is given by $\dfrac{f(\mathbf{x}_m \mid H_1)}{f(\mathbf{x}_m \mid H_0)}=\dfrac{f_T(-t_m \mid m-1,\delta\cdot\sqrt{m})}{f_T(t_m \mid m-1,\delta\cdot\sqrt{m})}$
where $t_m$ is the observed value of $T_m$ and $f_T(x \mid m-1,\delta\cdot\sqrt{m})$ is the p.d.f.\ at $x$ of the non-central T distribution with $m-1$ degrees of freedom and parameter $\delta\cdot\sqrt{m}$. 

\paragraph{$\Delta_{@t}^{I_1,I_0}.P_{\geq \theta}(\phi)$ and $\Delta_{@t}^{I_1,I_0}.R^{[a,b]}_{\geq \theta}$ formulas.} Since we can express the causal effect as the mean of a (non-Bernoulli) variable (the paired difference in the counterfactual outcomes of $I_1$ and $I_0$), we can apply the same SPRT procedure introduced above for $I_{@ t}.R_{\geq \theta}(C_{[a,b]})$ formulas
\footnote{For the special case of $\Delta_{@t}^{I_1,I_0}.P_{>0}(\phi)$, an alternative sequential test could be used, see~\cite{david2011statistical}.}, provided that we use the correct definition of $f(\mathbf{G}')$, i.e., that of~\eqref{eq:alt_effect_p} for $\Delta_{@t}^{I_1,I_0}.P_{\geq \theta}(\phi)$ and~\eqref{eq:alt_effect_r} for  $\Delta_{@t}^{I_1,I_0}.R^{[a,b]}_{\geq \theta}$. 

When $I_1=I_0$, however, the above procedure fails because the two policies attain the same outcomes, and so their pairwise differences are constantly $0$, resulting in $S_m=0$ and $T_m=\infty$, which has a likelihood of $0$. To detect this case, as done in~\cite{david2011statistical}, we run a dedicated SPRT to test that the probability of obtaining equal outcomes is equal to $1$.

\paragraph{Boolean combinations.} To verify $\neg \Phi$ with strength $\langle \alpha, \beta \rangle$, we verify $\Phi$ with strength $\langle \beta, \alpha \rangle$ and negate the result. To verify a conjunction $\bigwedge_{i=1}^N \Phi_i$ with strength $\langle \alpha, \beta \rangle$, we need to verify each conjunct $\Phi_i$ with strength $\langle \alpha/N, \beta \rangle$. See~\cite{younes2006statistical} for more details. 

\subsection{Quantitative Properties}
For quantitative properties, we use Chernoff-Hoeffding bounds to identify the number of realizations $n$ necessary such that the Monte-Carlo estimate of the probability (or reward) property meets \textit{a priori} error and confidence bounds. Given an error bound $\delta>0$ and an iid sample $X_1,\ldots, X_n$ such that for each $i=1,\ldots,n$, $\mathbb{E}[X_i]=\mu$ and $x_l\leq X_i \leq x_u$ for some constant $x_l<x_u$, then the Hoeffding inequality~\cite{hoeffding1963probability} establishes that
$P(| \bar{X}_n - \mu| \geq \delta)\leq 2\exp{\left(-\dfrac{2n\delta^2}{(x_u-x_l)^2}\right)},$
where $\bar{X}_n=(1/n) \sum_{i=1}^n X_i$ is the sample mean. 
Hence, given bounds $\delta>0$ and $0<\alpha<1$, one can determine \textit{a priori} the number of realizations $n$ such that $P(| \bar{X}_n - \mu| \geq \delta)\leq \alpha$, by equating $\alpha=2\exp{\left(-\dfrac{2n\delta^2}{(x_u-x_l)^2}\right)}$ and obtaining
$ n = \left\lceil - \dfrac{(x_u-x_l)^2\log{(\alpha/2)}}{2\delta^2} \right\rceil$. 

For the special case of $I_{@t}.P_{=?}(\phi)$ properties, $\bar{X}_n$ is the sample estimate of the probability, $\mu$ is the probability to estimate (and hence $0<\delta<1$), $x_l=0$ and $x_u=1$. For $\Delta_{@t}^{I_1,I_0}.P_{=?}(\phi)$ properties, we have that $x_l=-1$ and $x_u=1$ as these are the ranges for the difference of two Bernoulli outcomes. For $I_{@ t}.R^{[a,b]}_{=?}$ formulas, each realization is a cumulative reward value, hence $x_l=(b-a)R_{l}$ and $x_u=(b-a)R_{u}$ where $R_{u}$ and $R_{l}$ are, respectively, the largest and smallest values of the MDP's reward function $R$. Hence, for $\Delta_{@t}^{I_1,I_0}.R^{[a,b]}_{=?}$ formulas, we have $x_u=(b-a)(R_{u}-R_{l})$ and $x_l=(b-a)(R_{l}-R_{u})$.

These \textit{a priori} bounds, however, might be too conservative, especially for reward properties where the range $(x_u-x_l)$ tends to be consistently larger than what observed empirically. An alternative is to compute confidence intervals, i.e., fix the sample size $n$ and the confidence $1-\alpha$, thereby obtaining an estimate $\bar{X}_n$ and an interval $[\bar{X}_n]_{\alpha} \ni \bar{X}_n$ such that $P(\mu \not\in [\bar{X}_n]_{\alpha} )=\alpha$. In this sense, the width of $[\bar{X}_n]_{\alpha}$ is comparable to the $\delta$ bound in Hoeffding inequality. 
To construct confidence intervals for $I_{@t}.P_{=?}(\phi)$ properties, one can use the common normal-approximation (aka Wald) interval if $n$ is not too small or the true probability not too close to $0$ or $1$\footnote{For the normal approximation to be valid, we require $n\cdot p_{\phi}\geq 10$ and $n\cdot (1-p_{\phi})\geq 10$. }, or use the ``exact'' (but usually conservative) Clopper-Pearson interval. For the other properties, we can construct one-sample mean intervals using the T distribution.

%% file: sanity_check.tex
\section{MiniGrid Benchmark}\label{sec:sanity}
\begin{figure}
\centering
  \begin{minipage}[b]{0.3\textwidth}
    \includegraphics[width=\textwidth, trim={0 0 0 3.5cm},clip]{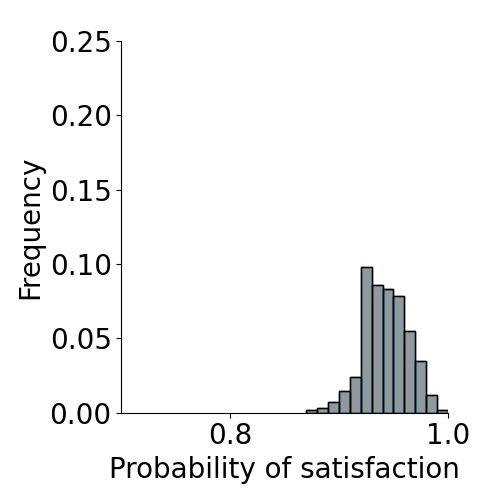}
  \end{minipage}\hspace{.5cm}
  \begin{minipage}[b]{0.55\textwidth}
       \caption{Sanity check experiment: comparison of post-interventional distribution (orange) and distribution of counterfactuals (blue). The two are indistinguishable, as desired.}
       \label{fig:from_start_optimal_vs_optimal}
  \end{minipage}
\end{figure}

In this section we provide an experiment as a sanity check for  counterfactual inference. We saw in Section~\ref{sec:scm_mdp_encoding} that Gumbel-Max  counterfactuals cannot be identified precisely, and so, we resort to approximate inference via rejection sampling. If the inference is accurate, we should see that the probability of $\phi$ on paths sampled directly from the post-interventional model $\gumbelscm{\mdp}[I]$ is equivalent to that on paths obtained by performing counterfactual inference on a set of nominal paths of $\gumbelscm{\mdp}$. Our results, reported in Fig.~\ref{fig:from_start_optimal_vs_optimal},  confirm this hypothesis (the two histograms are indistinguishable). 

\begin{figure}
    \centering
    \subfloat[]{\scalebox{0.6}{\input{imgs/from_start_optimal_vs_random.tex}}}\hspace{1cm}
    \subfloat[]{\scalebox{0.6}{\input{imgs/from_first_step_optimal_vs_random.tex}}}\hspace{1cm}
    \subfloat[]{\scalebox{0.6}{\input{imgs/from_first_step_optimal_vs_random_intrv.tex}}}
    \caption{Some example paths from the grid-world experiments. The robot's initial state is in the top-left cell, the fire represents the unsafe state, and the flag the goal state. The color code is the same as Fig.~\ref{fig:trajectories}. Black line: observed path,  dark blue: counterfactual path, light blue: future realization under $I$ (starting from counterfactual state), light grey: future realization under nominal policy (starting from observed state).}
    \label{fig:traces}
\end{figure}
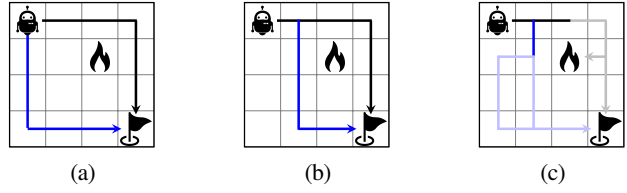

%% file: imgs/from_start_optimal_vs_random.tex
\begin{tikzpicture}[scale=.8]
	\draw[step=1cm,gray] (0,0) grid (4, 4);
	\node at (.5,3.5) {\includegraphics[scale=0.08]{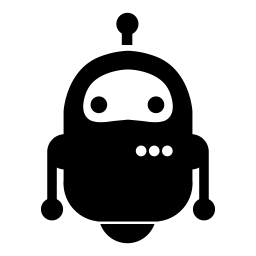}};
	\node at (2.5,2.5) {\includegraphics[scale=0.08]{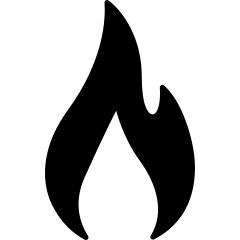}};
	\node at (3.5,.5) {\includegraphics[scale=0.01]{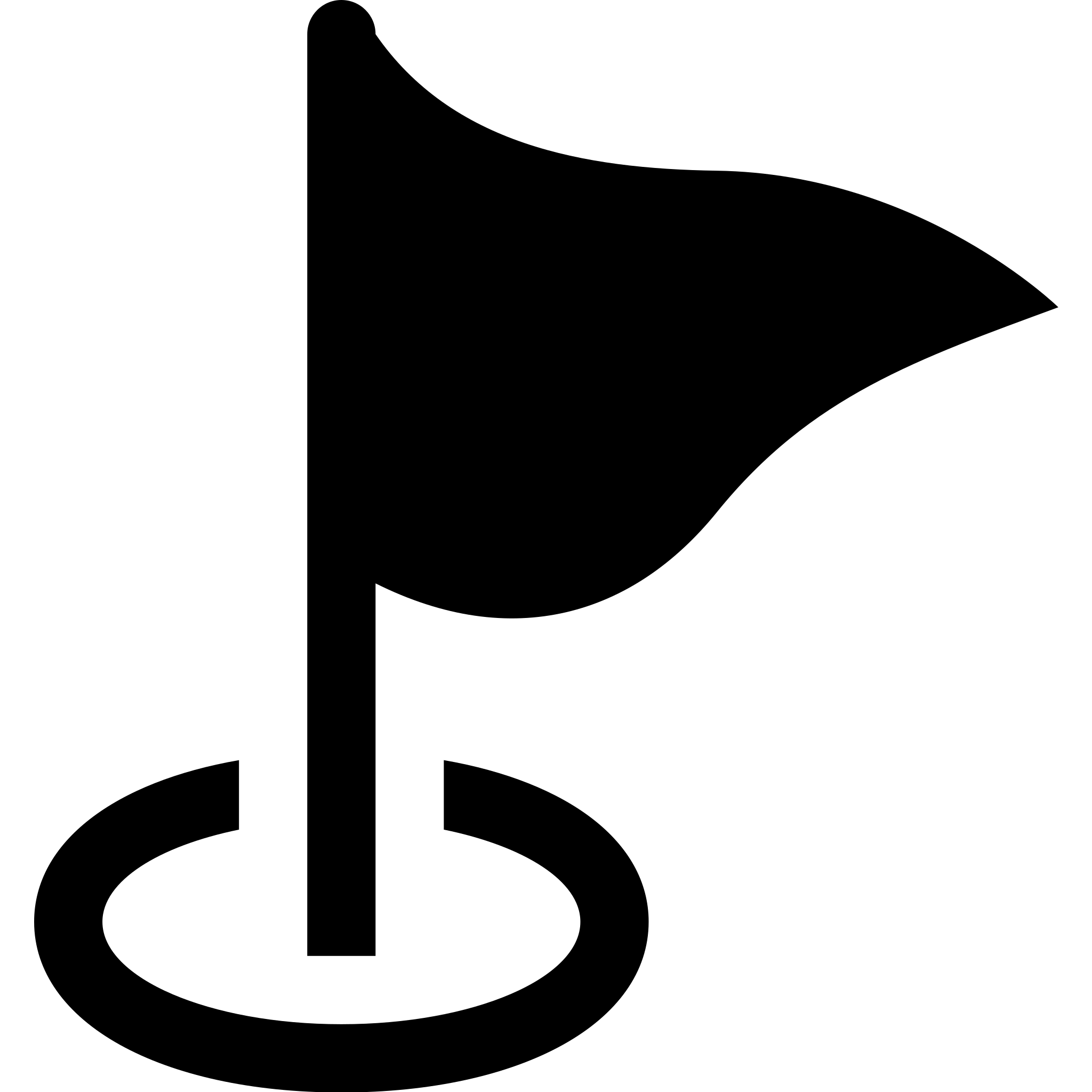}};
	\draw[thick] (0,0) rectangle (4,4);
	\draw[ultra thick, -, >=stealth, draw=black!100!white] (.9,3.5) -- (3.525,3.5);
	\draw[ultra thick, ->, >=stealth, draw=black!100!white] (3.5,3.5) -- (3.5,0.9);
	
	\draw[ultra thick, -, >=stealth, draw=blue!100!white] (.5,3.1) -- (.5,.525);
	\draw[ultra thick, ->, >=stealth, draw=blue!100!white] (.5,0.5) -- (3.1,0.5);

\end{tikzpicture}

%% file: imgs/from_first_step_optimal_vs_random.tex
\begin{tikzpicture}[scale=.8]
	\draw[step=1cm,gray] (0,0) grid (4, 4);
	\node at (.5,3.5) {\includegraphics[scale=0.08]{./imgs/roboto}};
	\node at (2.5,2.5) {\includegraphics[scale=0.08]{./imgs/flame}};
	\node at (3.5,.5) {\includegraphics[scale=0.01]{./imgs/flag}};
	\draw[thick] (0,0) rectangle (4,4);
	\draw[ultra thick, -, >=stealth, draw=black!100!white] (.9,3.5) -- (3.525,3.5);
	\draw[ultra thick, ->, >=stealth, draw=black!100!white] (3.5,3.5) -- (3.5,0.9);
	
	\draw[ultra thick, -, >=stealth, draw=blue!100!white] (1.5,3.525) -- (1.5,.475);
	\draw[ultra thick, ->, >=stealth, draw=blue!100!white] (1.5,0.5) -- (3.1,0.5);

\end{tikzpicture}

%% file: imgs/from_first_step_optimal_vs_random_intrv.tex
\begin{tikzpicture}[scale=.8]
	\draw[step=1cm,gray] (0,0) grid (4, 4);
	\node at (.5,3.5) {\includegraphics[scale=0.08]{./imgs/roboto}};
	\node at (2.5,2.5) {\includegraphics[scale=0.08]{./imgs/flame}};
	\node at (3.5,.5) {\includegraphics[scale=0.01]{./imgs/flag}};
	\draw[thick] (0,0) rectangle (4,4);
	\draw[ultra thick, -, >=stealth, draw=black!100!white] (.9,3.5) -- (2.525,3.5);
	\draw[ultra thick, -, >=stealth, draw=black!25!white] (2.525,3.5) -- (3.525,3.5);
	
	\draw[ultra thick, ->, >=stealth, draw=black!25!white] (3.5,3.5) -- (3.5,0.9);
    \draw[ultra thick, ->, >=stealth, draw=black!25!white] (3.5,2.5) -- (2.9,2.5);

	\draw[ultra thick, -, >=stealth, draw=blue!100!white] (1.5,3.525) -- (1.5,2.475);
	\draw[ultra thick, -, >=stealth, draw=blue!25!white] (1.5,2.525) -- (1.5,.475);
	\draw[ultra thick, ->, >=stealth, draw=blue!25!white] (1.5,0.5) -- (3.1,0.5);
     \draw[ultra thick, -, >=stealth, draw=blue!25!white] (.525,2.525) -- (.525,.475);
      \draw[ultra thick, -, >=stealth, draw=blue!25!white] (.5,2.5) -- (1.525,2.5);
    \draw[ultra thick, -, >=stealth, draw=blue!25!white] (.5,.5) -- (1.5,.5);

\end{tikzpicture}

%% file: bisimilarity.tex
\section{A discussion about finite-horizon bisimilarity}\label{sec:bisimilarity}
In this section, we show the MDP under policy $\pi$ and the corresponding Gumbel-max SCM are $T$-step bisimilar where $T$ is the horizon. 

We first define the notion of finite-horizon bisimulation as follows.

\np{

\begin{definition}[Finite-Horizon Bisimulation (adapted from~\cite{kamaleson2016finite})]\label{def:bisim}
Given an MDP $\mdp$, policy $\pi$, and time bound $T$, then
a \emph{$k$-step finite-horizon bisimulation} between the Gumbel-max SCM $\gumbelscm{\mdp,\pi,T}$ and the MDP $\mdp$ under policy $\pi$ is an equivalence relation $R_k \subseteq \states \times \states$, such that, for all states $(s^1, s^2) \in R_k$, the following two conditions are satisfied:

\begin{enumerate}
\item \label{first_bisim}$L(s^1) = L(s^2)$;
\item $P_{\gumbelscm{\mdp,\pi,T}}(S'\in C \mid S=s^1, A=\pi(s^1))=\sum_{s'\in C}\probs(s'\mid s^2, \pi(s^2))$, \ for all $ C \in \states/ R_{k-1},$

\end{enumerate}
where $\states/ R_{k-1}$ denotes the set of equivalence classes of set $\states$ by relation $R_{k-1}$ and $R_{k-1}$ is a ($k-1$)-step finite-horizon bisimulation. A $0$-step finite-horizon
bisimulation is an equivalence relation $R_0$ satisfying only condition \ref{first_bisim} above.

\end{definition}
Note that in the above definition we denote the SCM variables by $S$, $A$, and $S'$ instead of using the time-indexed notation $S_i$, $A_i$, and $S_{i+1}$,
because by the Markov property, conditional next state probabilities are not affected by the path position (and so, the choice of $i$ would have been purely arbitrary)\footnote{That is, $P_{\gumbelscm{\mdp,\pi,T}}(S_{i+1} \mid S_i, A_i) = P_{\gumbelscm{\mdp,\pi,T}}(S_{j+1} \mid S_j, A_j)$ for any $j$ and $i$ such that $S_i=S_j$ and $A_i=A_j$.}.
}

Now, we can define when we have finite-horizon bisimilarity. 
\begin{definition}[Finite-Horizon Bisimulation Equivalent~\cite{kamaleson2016finite}]
We say states $s^1$, $s^2$ are $(k\text{-step})$ finite-horizon bisimulation equivalent (bisimilar), denoted $s^1 \sim_k s^2$, if there exists a $k$-step finite-horizon bisimulation $R$ such that $(s^1, s^2) \in R$. Two states $s^1$ and $s^2$ satisfying $s^1 \sim_k s^2$ have the same stepwise behaviour over $k$ steps. 
\end{definition}
\np{We say that the MDP $\mdp$ under policy $\pi$ and the corresponding SCM $P_{\gumbelscm{\mdp,\pi,T}}$ are $(k\text{-step})$ finite-horizon bisimilar if $s \sim_k s$ for all states $s$ with $P_I(s)>0$ (or equivalently, with $P_{\gumbelscm{\mdp,\pi,T}}(S=s)>0$).}

Before showing that the MDP and its Gumbel-Max SCM translation are bisimilar, we briefly review the Gumbel-Max construction, as stated in \cite{huijben2022review}. 
Assume we have a categorical distribution $\operatorname{Cat(\boldsymbol{p})}$, where $\boldsymbol{p}= [p_1, \ldots, p_{|\states|}]^\top = \probs (s'\mid s, a)$ is the vector of probabilities of the $|\states|$ potential next states, given the current state $s$ and current action $a$. 
The Gumbel-Max trick is used to generate a sample from a categorical random variable $I \sim \operatorname{Cat(\boldsymbol{p})}$. To achieve this, independent and identically distributed (i.i.d.) Gumbel noise samples $(G_s)_{s\in \states}$ are added to the log probabilities $\log(p_s)$, and the index with the highest value is selected. The selected index follows a Gumbel distribution. To summarize, the Gumbel-max trick utilizes Gumbel noise and the selection of maximum value to obtain a sample from a categorical random variable. To put it formally:

\begin{equation}
\label{eq:gumbel_max_ap}
I = \underset{s \in \states}{\operatorname{argmax}} \ {(\log(p_s) + G_{s}) } \sim \operatorname{Cat}(\boldsymbol{p})
\end{equation}

The proof can be found in~\cite{huijben2022review}. \mk{As a direct conclusion from the Gumbel-max trick the probability distribution of the MDP under policy $\pi$ and the SCM are the same.}

\begin{equation}
\label{eq:gumbel_max_p}
P_{\gumbelscm{\mdp,\pi, T}}( S'= s'\mid S= s,  A= \pi(s)) = \probs(s'\mid s, \pi(s))
\end{equation}

Given the Gumbel-max trick and the notion of finite-horizon bisimilarity we can derive the following proposition.
\mk{
\begin{proposition}
Given an MDP $\mdp$, policy $\pi$, time bound $T$, 
then for any path $\tau$ of $\mdp$ induced by $\pi$ of length $T$
, the Gumbel-max SCM $\gumbelscm{\mdp,\pi,T}$ and $\mdp$ under policy $\pi$ are $T$-step finite-horizon bisimilar.
\end{proposition}
}
\begin{proof}
We aim to prove the Gumbel-max SCM $\gumbelscm{\mdp,\pi,T}$ and $\mdp$ under policy $\pi$ are $T$-step finite-horizon bisimilar. We accomplish this by defining the following relation $R_k= \{(s, s)\mid s\in \states\}$ for all $k\leq T$ and showing that $R_k$ is a bisimulation as per Definition~\ref{def:bisim}.

Initially, we aim to illustrate that the $\gumbelscm{\mdp,\pi,T}$ and $\mdp$ are $0$-step bisimilar. From the definition of $R_k$, it's clear that the states of $\gumbelscm{\mdp,\pi,T}$ and $\mdp$ share the same label. So, the first condition is naturally satisfied for all states. This condition alone is sufficient for $0$-step bisimilarity, thereby we conclude $\gumbelscm{\mdp,\pi,T}$ and $\mdp$ under policy $\pi$ are $0$-step bisimilar.

Our next objective is to establish $\gumbelscm{\mdp,\pi,T}$ and $\mdp$ are $k$-step bisimilar. As we mentioned before, the first condition is trivially satisfied. We now turn our focus to the second condition. Given the above construction for $R_k$, it is easy to see that the equivalence class of any state $s$ w.r.t.\ $R_k$ (for any $k$) is the singleton $\{s\}$ and thus, the quotient $\states/ R_{k-1}$ corresponds to the partition $\{\{s\} \mid s \in \states\}$. 
This makes condition 2 of Definition~\ref{def:bisim} equivalent to Eq.~\ref{eq:gumbel_max_p} for all $k\leq T$.
Hence, the SCM and the MDP under policy $\pi$ are indeed $k$-step bisimilar. Since the above reasoning holds  for all $k\leq T$, then the SCM and the MDP under policy $\pi$ are $T$-step bisimilar.

\end{proof}